\documentclass[11pt]{article}

\usepackage[preprint]{acl}

\usepackage{times}
\usepackage{latexsym}

\usepackage[T1]{fontenc}

\usepackage[utf8]{inputenc}

\usepackage{microtype}

\usepackage{inconsolata}

\usepackage{amsmath}
\usepackage{amssymb}
\usepackage{mathtools}
\usepackage{amsthm}

\usepackage[capitalize,noabbrev]{cleveref}

\usepackage{url}            
\usepackage{booktabs}       
\usepackage{amsfonts}       
\usepackage{nicefrac}       
\usepackage{xcolor}         

\usepackage{enumitem}
\usepackage{graphicx}
\usepackage{subfigure}
\usepackage{multirow}
\usepackage{tabularx}
\usepackage{wrapfig}

\usepackage{tcolorbox}
\tcbuselibrary{breakable}

\usepackage[disable,textsize=tiny]{todonotes}

\newtheorem{theorem}{Theorem}

\newtheorem{definition}[theorem]{Definition}

\title{Enhancing Reinforcement Learning Fine-Tuning with an Online Refiner}

\author{Hao Ma\textsuperscript{1,2} \and
Zhiqiang Pu\textsuperscript{1,2}\thanks{Corresponding author: zhiqiang.pu@ia.ac.cn} \and
Yang Liu\textsuperscript{1,2} \and
Xiaolin Ai\textsuperscript{2} \\
\textsuperscript{1}School of Artificial Intelligence, University of Chinese Academy of Sciences \\
\textsuperscript{2}Institute of Automation, Chinese Academy of Sciences \\
\texttt{\{mahao2021, zhiqiang.pu, liuyang2025, xiaolin.ai\}@ia.ac.cn}}

\begin{document}
\maketitle

\begin{abstract}
    Constraints are essential for stabilizing reinforcement learning fine-tuning (RFT) and preventing degenerate outputs, yet they inherently conflict with the optimization objective because stronger constraints limit the ability of a fine-tuned model to discover better solutions. We propose \textit{dynamic constraints} that resolve this tension by adapting to the evolving capabilities of the fine-tuned model based on the insight that constraints should only intervene when degenerate outputs occur. We implement this by using a reference model as an \textit{online refiner} that takes the response from the fine-tuned model and generates a minimally corrected version which preserves correct content verbatim while fixing errors. A supervised fine-tuning loss then trains the fine-tuned model to produce the refined output. This mechanism yields a constraint that automatically strengthens or relaxes based on output quality. Experiments on dialogue and code generation show that dynamic constraints outperform both KL regularization and unconstrained baselines, achieving substantially higher task rewards while maintaining training stability.
\end{abstract}

\section{Introduction}

Reinforcement learning fine-tuning (RFT) has emerged as a powerful paradigm for aligning large language models (LLMs) with task-specific objectives, achieving remarkable success across diverse domains including safety alignment \citep{ji2023beavertails, liu2023importance}, code generation \citep{gu2023llm, shojaee2023execution}, and reasoning \citep{guo2025deepseek, team2025kimi}. Recent algorithmic innovations, such as GRPO~\citep{shao2024deepseekmath}, DAPO~\citep{yu2025dapo}, VAPO~\citep{yuan2025vapo}, and GSPO~\citep{zheng2025group}, have continually pushed the performance boundaries of RFT. However, as reinforcement learning (RL) plays an increasingly prominent role in post-training, it suffers from catastrophic forgetting~\citep{chen2025beyond, korbak2022rl}.

This phenomenon arises from the inherent difficulty of tuning a model in a high-dimensional exploration space with sparse rewards. The RL algorithm selectively reinforces task-specific knowledge, causing the policy distribution to become increasingly sharp and concentrated~\citep{walder2025passk}. As the fine-tuned policy \(\pi_\theta\) deviates substantially from the reference policy \(\pi_0\), the general capabilities acquired during pretraining and supervised fine-tuning (SFT) are gradually eroded~\citep{chen2025beyond}. In extreme cases, this catastrophic forgetting can severely impair the model's expressive capacity, leading to distribution collapse where the model degenerates into generating incoherent or nonsensical outputs~\citep{korbak2022rl,ma2024coevolving,liu2025scaling, mai2025agent}.

A natural mitigation strategy is to impose a regularization term that constrains the fine-tuned policy from drifting too far from the reference one. KL divergence regularization, widely adopted in early RFT algorithms~\citep{korbak2022rl, padula2024exploring}, serves precisely this purpose by anchoring \(\pi_\theta\) to \(\pi_0\). However, this approach introduces a fundamental trade-off. While it prevents catastrophic forgetting, it also limits \(\pi_\theta\) to stay close to \(\pi_0\), inherently restricting exploration and preventing the discovery of optimal policies that may reside in distant regions of the solution space~\citep{yu2025dapo}.

Our key insight is that if we pursue extreme performance on specific tasks, we should abandon KL regularization, yet an alternative constraint remains essential to prevent the incoherent outputs induced by catastrophic forgetting. The fundamental limitation of KL regularization lies in its property of being a \emph{static constraint}. In this paper, we propose replacing this static constraint with a \emph{dynamic constraint} that adapts to the evolving capabilities of the fine-tuned policy. Rather than anchoring to a fixed reference, the dynamic constraint guides \(\pi_\theta\) toward an adaptive target that progresses alongside the policy.

\begin{figure*}[thp]
    \centering
    \subfigure[Static constraint]{\includegraphics[width=0.4\linewidth]{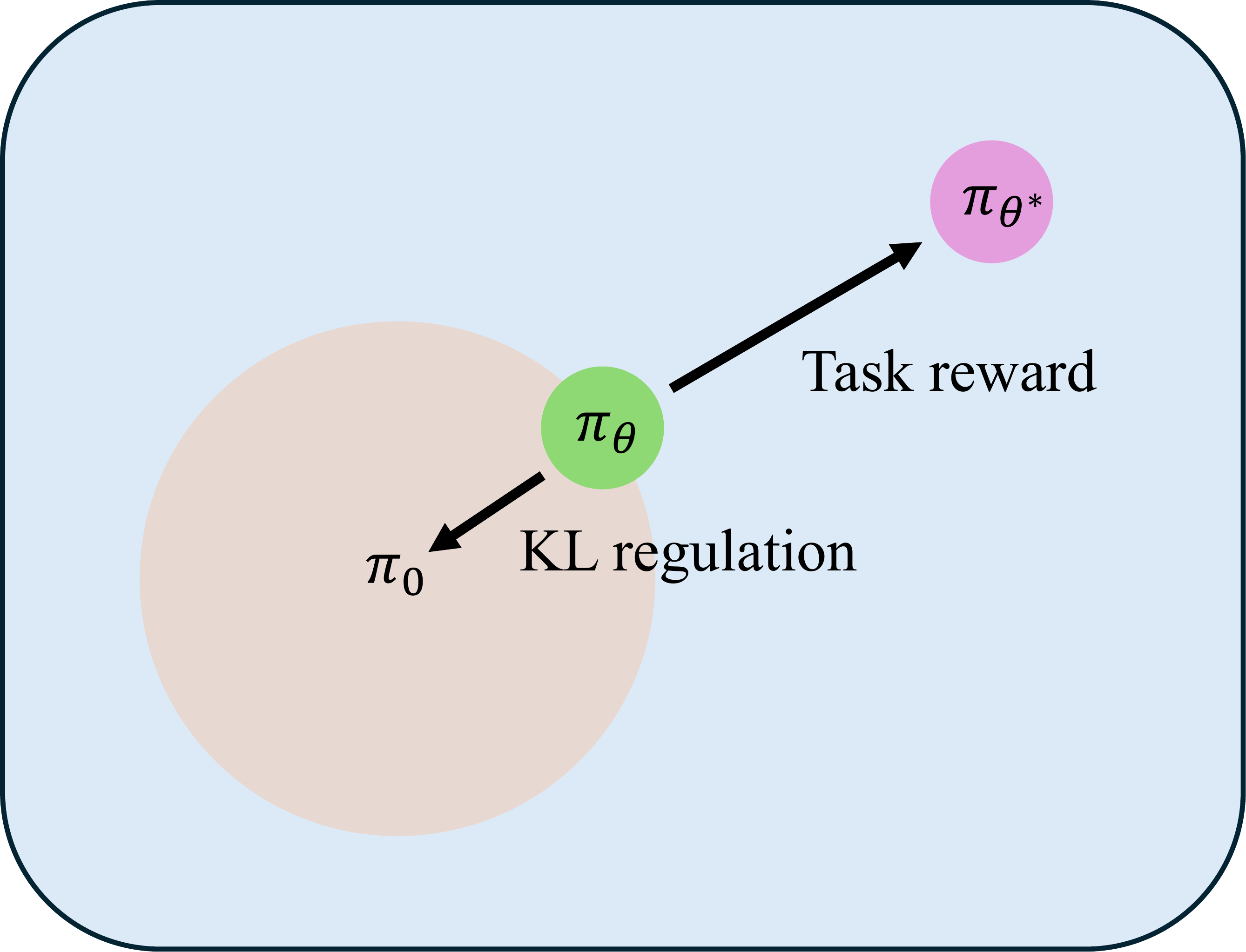}}
    \hspace{0.5cm}
    \subfigure[Dynamic constraint]{\includegraphics[width=0.4\linewidth]{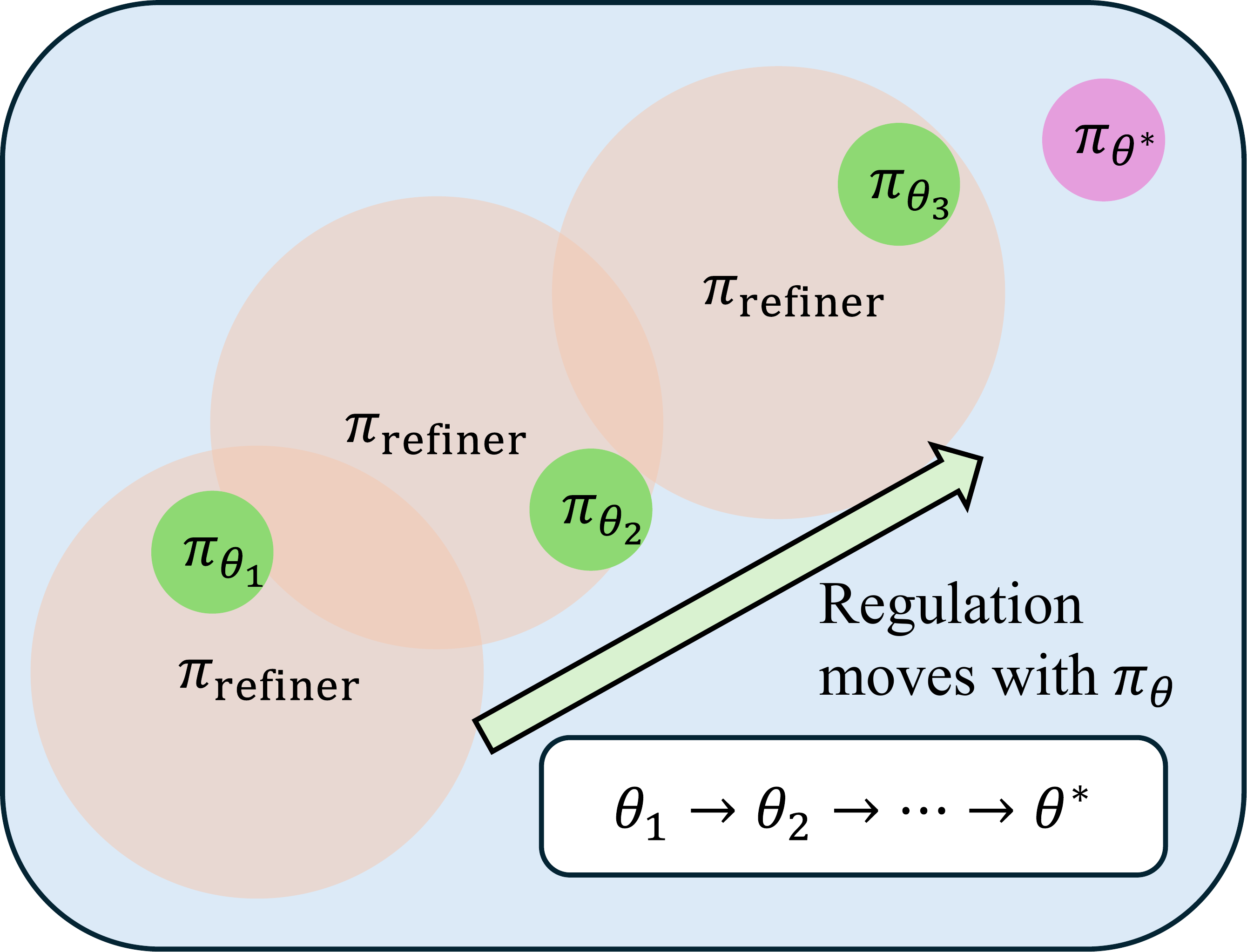}}
    \caption{\textbf{An illustration of the insight of dynamic constraint.}
    (a) The left figure illustrates the conventional KL regularization, which constrains the fine-tuned policy $\pi_{\theta}$ to remain close to the reference policy $\pi_{0}$. However, when the optimal policy lies far from $\pi_{0}$ and requires $\pi_{\theta}$ to deviate significantly, the KL regularization becomes an obstacle to policy optimization.
    (b) The right figure presents the insight of the dynamic constraint, where the constraint is derived from $\pi_{0}$’s refined response based on $\pi_{\theta}$'s response and context. When $\pi_{\theta}$ deviates substantially from $\pi_{0}$, the dynamic constraint not only avoids hindering policy optimization but also provides effective guidance and correction for $\pi_{\theta}$.}

    \label{fig:graph_abs}
\end{figure*}

We realize this idea by prompting the reference model \(\pi_0\) to act as an \emph{online refiner}. Given an input query and the response from \(\pi_\theta\), the refiner generates a refined version of the response. The dynamic constraint is then defined as the cross-entropy teaching \(\pi_\theta\) to produce the refined response. This design maximally relaxes the constraint on \(\pi_\theta\) by only intervening when degenerate outputs occur. When \(\pi_\theta\) produces a coherent response, the refiner reproduces it verbatim and imposes almose no constraint. Only when degenerate outputs occur does the refiner apply targeted and minimal corrections. As illustrated in Figure~\ref{fig:graph_abs}, while the static constraint pulls the policy backward toward a fixed reference, the dynamic constraint propels it forward toward an adaptive, higher-quality target. From a broader perspective, our approach unifies RL with SFT on a continuously updated dataset.

We validate our approach through comprehensive experiments on dialogue and code generation tasks. We first integrate the dynamic constraint into PPO to analyze its training dynamics compared with static KL regularization, then incorporate it into DAPO~\citep{yu2025dapo}, a state-of-the-art RFT algorithm, and evaluate on challenging code generation benchmarks. Our results reveal two compelling advantages. First, the dynamic constraint substantially stabilizes training by reducing variance and preventing distribution collapse. Second, it enables policies to achieve significantly higher rewards by eliminating the performance ceiling imposed by static constraints. These findings underscore the potential of dynamic constraints for advancing RFT. More broadly, our approach establishes a self-elevating cycle between RL and SFT, opening new avenues for improving both paradigms.

\section{Preliminary}
\subsection{Problem Formulation}
To describe RFT using standard RL terminology, we formulate next-token prediction in causal language models as a sequential decision-making process. This is formally modeled as a language-augmented Markov Decision Process \citep{li2022pre}, defined as \( \mathcal{M} = \langle \mathcal{V}, \mathcal{S}, \mathcal{A}, r, P, \gamma \rangle \). Here, \( \mathcal{V} \) denotes the vocabulary of the language model, consisting of all possible tokens, and \( w \in \mathcal{V} \) refers to a token. The state space is defined as \( \mathcal{S} \subset \mathcal{V}^M \), where \( \mathcal{V}^M \) represents the set of all token sequences of length up to \( M \). Similarly, the action space is \( \mathcal{A} \subset \mathcal{V}^N \), where \( \mathcal{V}^N \) denotes token sequences of length up to \( N \). Here, \( M \) and \( N \) are the maximum token lengths for states and actions, respectively. 
A state \( s \in \mathcal{S} \) is represented as a token sequence \( s = (w_1, w_2, \dots, w_M) \), and an action \( a \in \mathcal{A} \) is defined as a generated sequence \( a = (w_1, w_2, \dots, w_N) \). For simpliciy, we denote $(w_1,\dots,w_i,\dots,w_{t-1})$ as $a_{<t}$, and $w_i$ as $a_i$. If the sequence length is shorter than the maximum, it is padded with a special token. The reward function \( r: \mathcal{S} \times \mathcal{A} \rightarrow \mathbb{R} \) assigns a task-specific scalar reward to each state-action pair. The transition function \( P: \mathcal{S} \times \mathcal{V} \rightarrow \mathcal{S} \) defines a deterministic transition based on autoregressive generation. At each timestep, the predicted token is appended to the current state: \(s_{i} = (s_{i-1}, a_{i}) = (s_0, a_{<i+1})\), where \( s_0 \) is the tokenized user input and $a_{<i+1}$ is the partial output sequence. The token-level policy of the language model is denoted by \( \pi(a_i \mid s_0, a_{<i}) \), and the sentence-level policy is defined as the product of token-level predictions: $\pi(a|s_0)=\prod_{i=1}^{N}\pi(a_i|s_0,a_{<i})$. The task reward \( r(s_0, a) \) is only available after the full sequence \( a \) is generated.

\subsection{Mirror Learning}
\label{sec:mirror_learning}
Mirror learning \citep{grudzien2022mirror} provides a unified theoretical framework for modern RL algorithms, including TRPO~\citep{schulman2015trust} and PPO~\citep{schulman2017proximal}. Policy optimization under mirror learning is expressed as:
\begin{equation}
    \label{eq:mirror_learning_update}
    \begin{split}
      \pi_{\text {new}} = \underset{\bar{\pi} \in \mathcal{N}(\pi_{\text {old}})}{\arg \max } \mathbb{E}_{s \sim \beta_{\pi_{\text {old}}} , a \sim \bar\pi} \big[ & A_{\pi_{\text {old}}}(s, a) \\
      & - \mathfrak{D}_{\pi_{\text {old}}}(\bar{\pi} \mid s) \big],
    \end{split}
\end{equation}
where $\mathcal{N}(\pi_{\text {old}})$ denotes the neighborhood of $\pi_{\text {old}}$, and $\mathfrak{D}_{\pi_{\text {old}}}(\bar{\pi} \mid s)$ is a drift term that quantifies the divergence between $\pi_{\text{old}}$ and a candidate policy $\bar\pi$. This framework guarantees convergence to the optimal policy if the operator $\mathcal{N}$ and $\mathfrak{D}$ satisfy mild requirements (see Appendix~\ref{appendix:proof} for details).
Modern RFT algorithms, such as PPO, GRPO, and DAPO, can all be viewed as instantiations of this general form, where the drift term controls update steps and the neighborhood defines the feasible search region.

\section{Method}
\label{sec:method}

\begin{figure*}[thbp]
    \centering
    \includegraphics[width=0.65\linewidth]{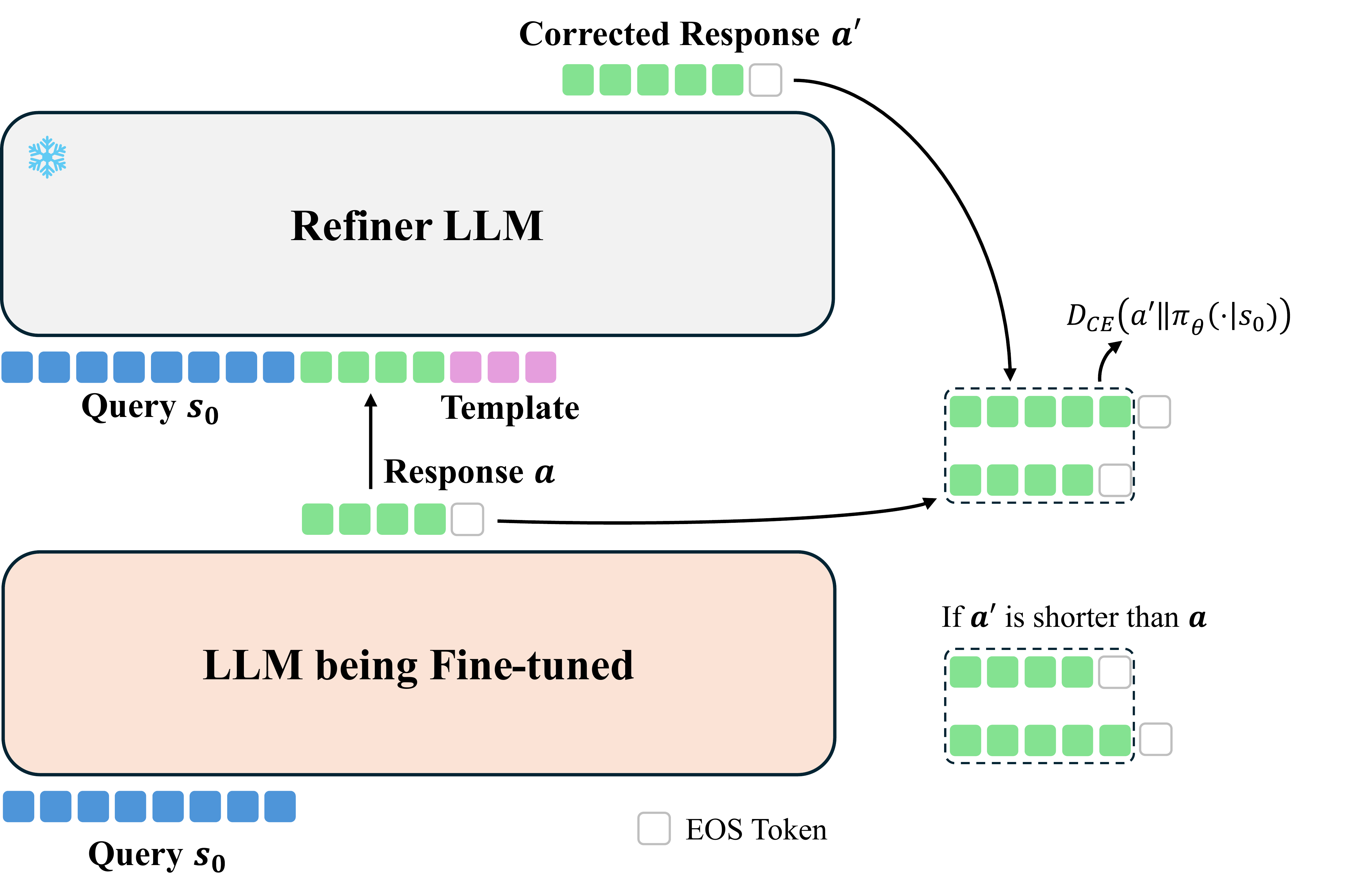}
    \caption{\textbf{The pipeline for calculating a dynamic constraint.} After $\pi_\theta$ generates a response $a$ for the query $s_0$, a refiner LLM is employed to refine the response based on the necessary context (quesry $s_0$, response $a$) and a predefined template. The refinement process is conservative. In most cases, the refiner simply repeats $a$ when no obvious issues are detected, while in other cases it makes only minimal edits to $a$ when necessary. The detailed template is provided in Appendix~\ref{appendix:exp_details}. The dynamic constraint is computed as a cross entropy loss that treats the refined response $a'$ as the ground truth, which is equivalent to the SFT loss on the pair $\langle s_0, a' \rangle$.}
    \label{fig:framework}
\end{figure*}

\subsection{RFT under a Static Constraint}
Current RL algorithms for fine-tuning LLMs typically include both task reward and KL regularization. The KL regularization is a constraint term that is transformed into an optimization objective. By treating the KL regularization as a constraint, the PPO algorithm can be viewed as solving the constrained optimization problem shown below.
\begin{equation}
\begin{split}
    \pi_{\text {new }}
    &=\underset{\bar{\pi} \in \mathcal{N}(\pi_0)}{\arg \max } \mathbb{E}_{s_0 \sim D, \{a_i\}_{i=1}^t \sim \pi_{\text{old}}}\\
    &\left[\min \left(\mathrm{r}(\bar{\pi})A_{\pi_{\text {old }},t}, \operatorname{clip}(\mathrm{r}(\bar{\pi}), 1 \pm \epsilon)A_{\pi_{\text {old }},t}\right)\right],
\end{split}
\end{equation}

\begin{equation}
    \mathcal{N}(\pi_0)=\left\{\bar{\pi} \in \Pi \mid D_{\mathrm{KL}}(\bar{\pi} \mid \pi_0) \leq \delta\right\},
\end{equation}
where we denote $A_{\pi_{\text {old }}}(s_{0}, a_{<t},a_t)$ as $A_{\pi_{\text {old}},t}$ for simplicity. The KL regularization constrains \(\pi_\theta\) to remain close to the reference model \(\pi_0\). This restriction can prevent \(\pi_\theta\) from reaching the optimal policy \(\pi_{\theta^*}\) and disrupts the convergence guarantee of standard PPO (Appendix~\ref{appendix:proof}). 

If the deviation between the fine-tuned model and the reference model is small, the KL regularization may have little influence on learning dynamics \citep{bai2022training}. However, when a larger deviation is desired, the KL regularization becomes a constraint that hinders progress. Since the KL divergence is measured relative to a fixed model, it restricts the direction of policy updates. This observation is consistent with recent work on reasoning models \citep{guo2025deepseek, team2025kimi, abdin2025phi} which suggests that reasoning capabilities depend on starting from a strong model.

Despite its limitations, removing the KL regularization is often unwise. RL for LLMs operates in an extremely large action space and faces sparse rewards. Without any form of constraint, policy optimization often leads to incoherent or meaningless outputs. This issue is known as distribution collapse \citep{korbak2022rl, ma2024coevolving}.

These factors create the KL regularization challenge. The KL term helps prevent incoherent outputs but also limits how far the policy can move from the reference model. The main task is to allow the model to explore different behaviors without causing distribution collapse.

\subsection{RFT under a Dynamic Constraint}
\label{sec:method-dynamic}
To address this challenge, we propose a dynamic constraint that provides timely corrections while adapting as \(\pi_\theta\) changes. This approach is motivated by the concept of a reference policy that stays near the current policy but avoids degenerate behaviors such as hallucinations. Since such a policy is not directly available, we approximate it by using the in-context learning capabilities of the reference model \(\pi_0\).

As illustrated in Figure~\ref{fig:framework}, we first generate a response \(a\) from \(\pi_\theta\) given query \(s_0\). Then the query \(s_0\) and response \(a\) are provided as input to the reference model \(\pi_0\) to produce an improved response \(a'\). This refined response serves as a dynamic anchor to guide the learning process. The dynamic constraint provides structure-preserving regularization without overly restricting policy improvement. 

\begin{equation}
\begin{split}
    \pi_{\text {new}}
    &=\underset{\bar{\pi} \in \mathcal{N}(\pi_{\text{refiner}})}{\arg \max } \mathbb{E}_{s_0 \sim D, \{a_i\}_{i=1}^t \sim \pi_{\text{old}}} \\
    &\left[\min \left(\mathrm{r}(\bar{\pi})A_{\pi_{\text {old }},t}, \operatorname{clip}(\mathrm{r}(\bar{\pi}), 1 \pm \epsilon)A_{\pi_{\text {old }},t}\right)\right],
\end{split}
\end{equation}

\begin{equation}
    \mathcal{N}(\pi_{\text{refiner}})=\left\{\bar{\pi} \in \Pi \mid D(\bar{\pi} \mid \pi_{\text{refiner}}) \leq \delta\right\}.
\end{equation}

In this formulation, \(\pi_{\text{refiner}}\) denotes the refiner output \(\pi_0(\cdot \mid s_0, a)\) where \(a\) is sampled from \(\pi_\theta\). Although the refiner can correct errors while preserving correct parts of the output, we notice that for the preserved tokens, $\pi_{\text{refiner}}$ can at best ensure that $arg\max\ \pi_0(\cdot|s_i, a) = arg\max\ \pi_\theta(\cdot|s_i)$, but it does not guarantee that $\pi_0(\cdot|s_i, a) = \pi_\theta(\cdot|s_i)$. So, directly applying a KL divergence constraint of the form $D_{\mathrm{KL}}(\pi_0(\cdot \mid s_i, a), \bar{\pi}(\cdot \mid s_i))$
is not appropriate. This is because \( \pi_0(\cdot \mid s_i, a) \) is only guaranteed to be more reliable than \( \bar{\pi}(\cdot \mid s_i) \) at the token level, and does not necessarily represent a better distribution overall.

\subsection{The Implementation of Dynamic Constraint}

\begin{figure}[btp]
    \centering
    \includegraphics[width=0.9\linewidth]{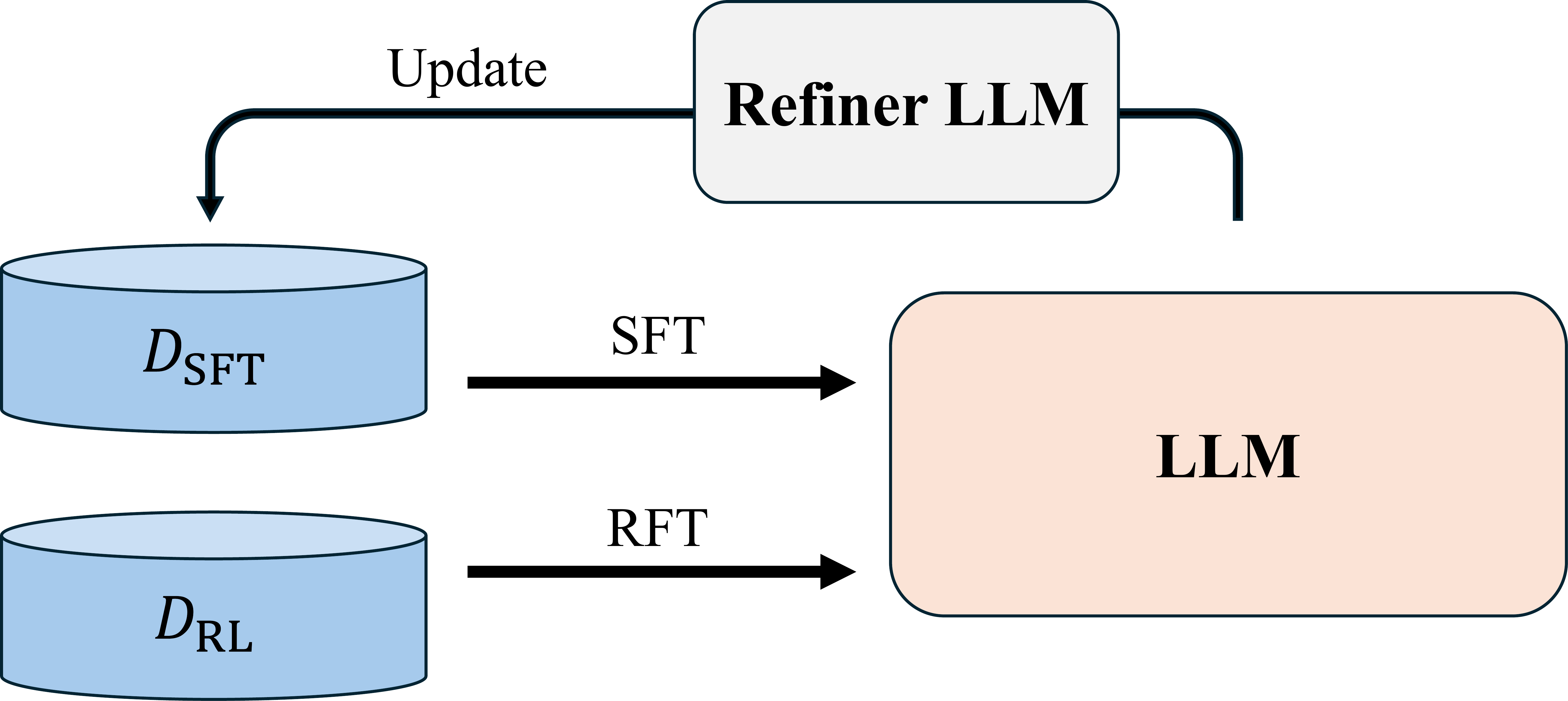}
    \caption{\textbf{Dynamic constraint from a dataset perspective.} The dynamic constraint can be interpreted as an RFT-SFT hybrid training approach with a dynamically updated SFT dataset.}
    \label{fig:data_perspective}
\end{figure}

We define the dynamic constraint in the form of a sequence-level cross entropy under self-feeding:

\begin{equation}
\begin{aligned}
    D(\bar{\pi} | \pi_{\text{refiner}})
    &=\! \sum_{t=1}^{\min\{|a|,|a'|\}} \!\!\!\!\! -\log\big(\pi_\theta(a'_t \mid s_0, a'_{<t})\big), \\
    &\text{s.t. } r(s_0, a') > r(s_0, a),
\end{aligned}
\label{eq:ce}
\end{equation}

where $a$ is the response sampled from $\pi_\theta$ given $s_0$, $a'$ is the refined response conditioned on $(s_0, a)$, and $r(\cdot,\cdot)$ denotes the reward function. In practice, we implement the dynamic constraint as an additive penalty: at step $t$, we add $-\eta \log\big(\pi_\theta(a'_t \mid s_0, a'_{<t})\big)$ on the reward, where $\eta$ is a penalty weight. This effectively integrates an SFT-style loss into the RL objective. The refined answer \(a'\), paired with the question \(s_0\), forms an SFT training example \(\langle s_0, a' \rangle\). Unlike prior RFT--SFT mixed training methods \citep{lv2025unified, liu2025uft, huang2025prefixrft}, our SFT data come from a dynamic dataset (Fig.~\ref{fig:data_perspective}), in which \(a'\) is distributed around \(a\). This prevents \(\pi_\theta\) from being confined to the neighborhood of a fixed reference policy.

Ideally, since the refiner can always choose to repeat \(a\) and only refine when \(a\) is problematic, the refined output \(a'\) should never be worse than \(a\). In practice, however, the refiner is imperfect, and cases where \(a'\) is worse than \(a\) are unavoidable. Moreover, as the capability of \(\pi_\theta\) improves during training, the proportion of such worse \(a'\) tends to increase. To prevent these inferior \(a'\) from negatively affecting the training of \(\pi_\theta\), we filter them out and exclude them from providing constraints.

\section{Experiment}
\label{sec:exp}

Our experimental evaluation is designed to answer two fundamental questions: (1) \textbf{Dynamics}: Can dynamic constraints effectively resolve the tension between training stability and unbounded exploration? (2) \textbf{Performance}: Does this mechanism yield state-of-the-art results on complex tasks compared to strong RFT baselines?

\subsection{Analysis on Training Dynamics}
\label{sec:exp_part1}
\textbf{Setup.} We investigate whether dynamic constraints can resolve the fundamental trade-off between training stability and exploration capability. We compare our \textit{Dynamic} approach against two key baselines: \textit{Static}, which employs standard KL regularization anchored to a fixed pretrained model, and \textit{w/o Constraint}, which performs pure RL optimization without any regularization. All constraint variants are implemented using PPO as the underlying RL algorithm.

The experiments span two benchmarks, including Prompt-Collection-v0.1 \citep{openrlhf2024prompt}, and APPS \citep{hendrycksapps2021}. The Prompt-Collection-v0.1 dataset consists of 179,000 QA samples for training, including prompts from six subsets: UltraFeedback \citep{cui2024ultrafeedback}, HelpSteer \citep{wang2024helpsteer}, OpenOrca Pairs \citep{openorca2023}, UltraInteract \citep{yuan2024ultrainteract}, DIBT 10K Prompts Ranked, and Capybara Preferences \citep{capybara_preferences}. We employ a reward model\footnote{\url{https://huggingface.co/OpenRLHF/Llama-3-8b-rm-mixture}} to provide the rewards. The APPS dataset comprises 10,000 problems sourced from open-access coding platforms such as Codeforces, Kattis, and others, with 5,000 problems for training and 5,000 for testing. This dataset is designed to evaluate the ability of language models to generate code from natural language specifications. Our reward model follows the framework of \citep{le2022coderl} and \citep{liu2023rltf}, utilizing a verifiable reward function based on compile and unit test results.

We employ Llama-3.2-3B-Instruct\footnote{\url{https://huggingface.co/meta-llama/Llama-3.2-3B-Instruct}} as our base model for training on both Prompt-Collection-v0.1 and APPS datasets. For comprehensive analysis, we report training curves for task rewards and KL divergence between the reference policy $\pi_0$ and the current policy $\pi_\theta$ across all benchmarks. Note that while the \textit{Dynamic} method does not explicitly use KL divergence as a constraint, we compute it for comparative analysis alongside the \textit{w/o Constraint} baseline. Both \textit{Dynamic} and \textit{Static} methods use Llama-3.2-3B-Instruct as their reference model.

\begin{figure*}[t!hb]
  \centering
  \subfigure[Task reward]{\includegraphics[width=0.3\linewidth]{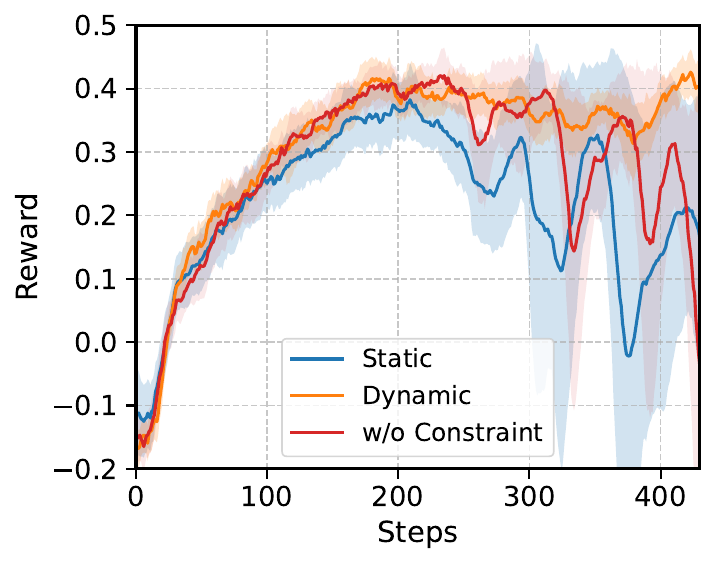}}  
  \hspace{0.1cm}
  \subfigure[KL divergence]{\includegraphics[width=0.3\linewidth]{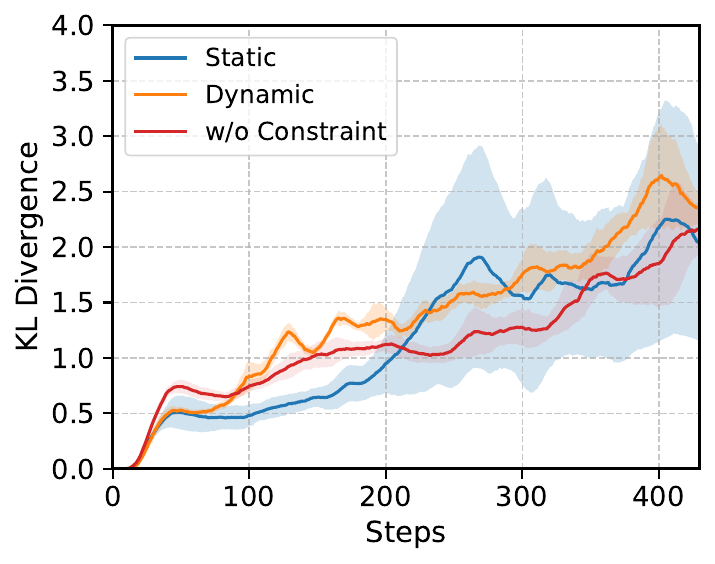}}
  \hspace{0.1cm}
  \subfigure[Cross entropy]{\includegraphics[width=0.3\linewidth]{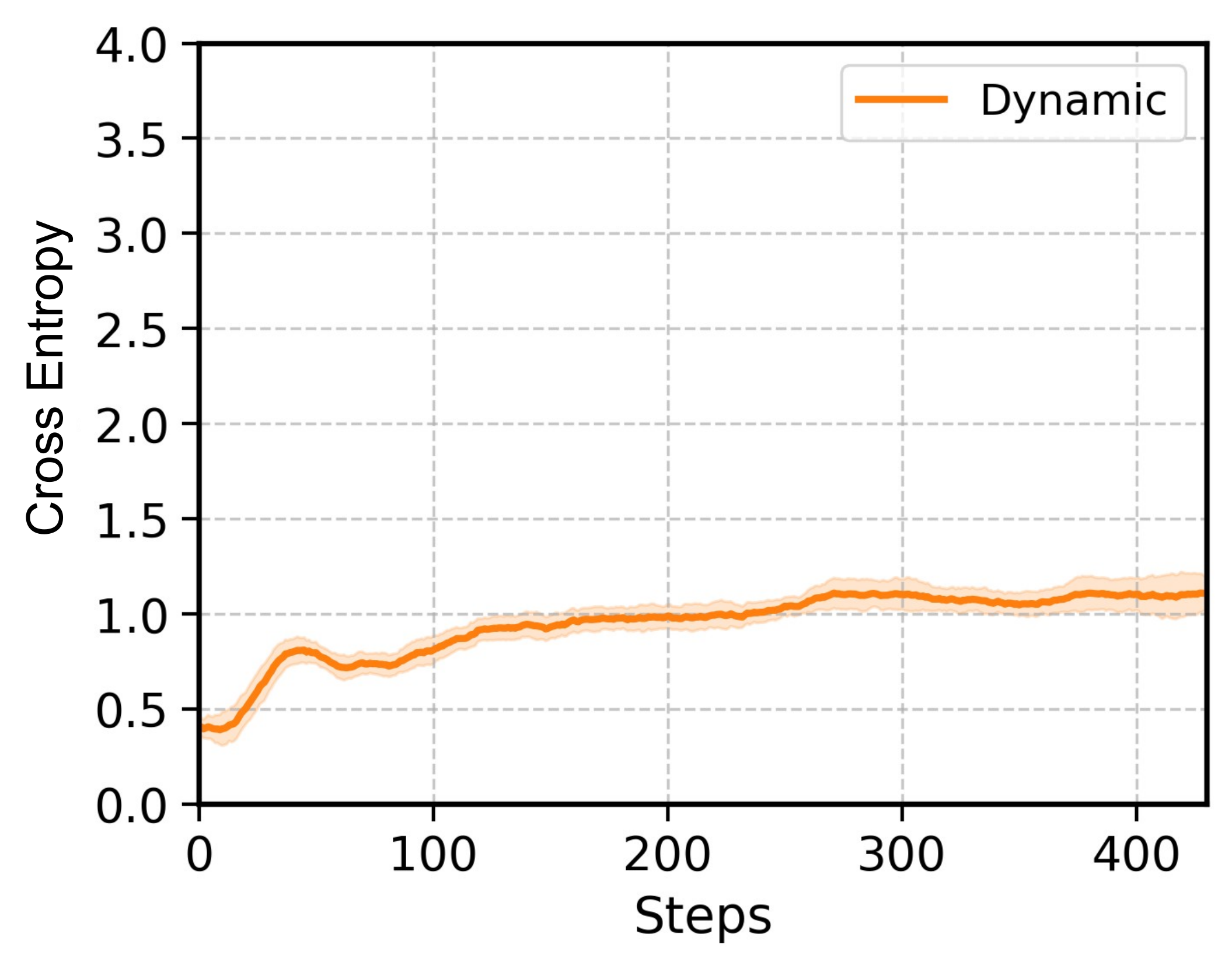}}  

  \subfigure[Task reward]{\includegraphics[width=0.31\linewidth]{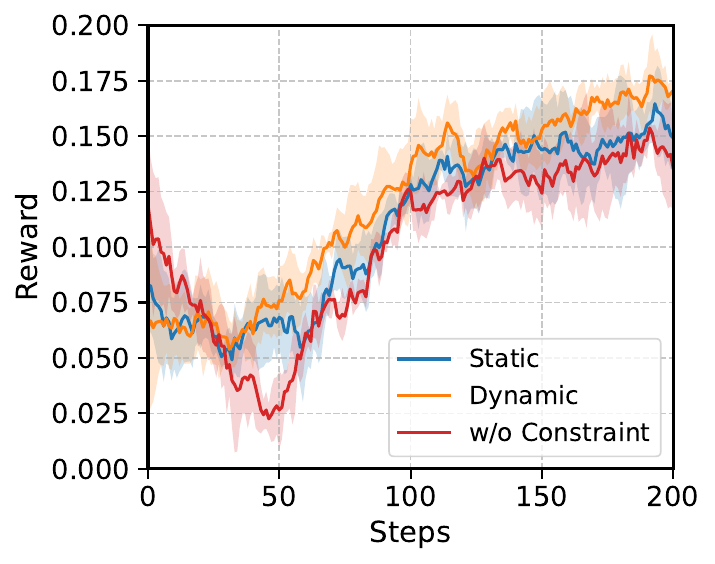}}  
  \subfigure[KL divergence]{\includegraphics[width=0.31\linewidth]{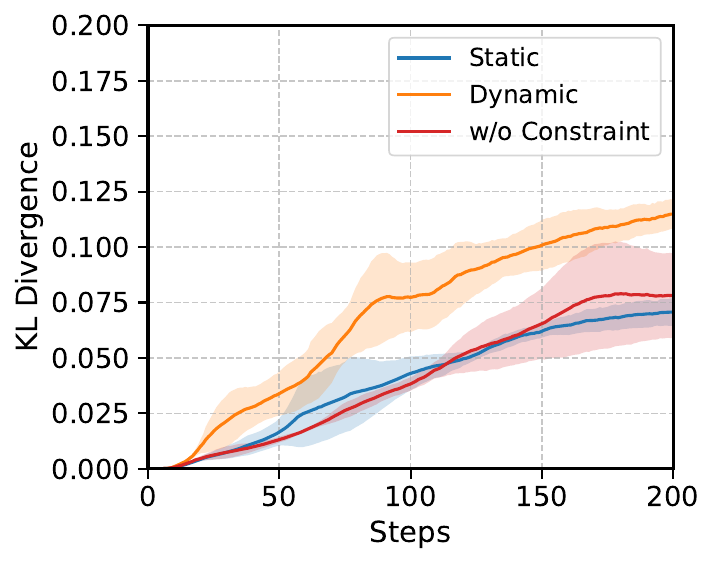}}  
  \subfigure[Cross entropy]{\includegraphics[width=0.31\linewidth]{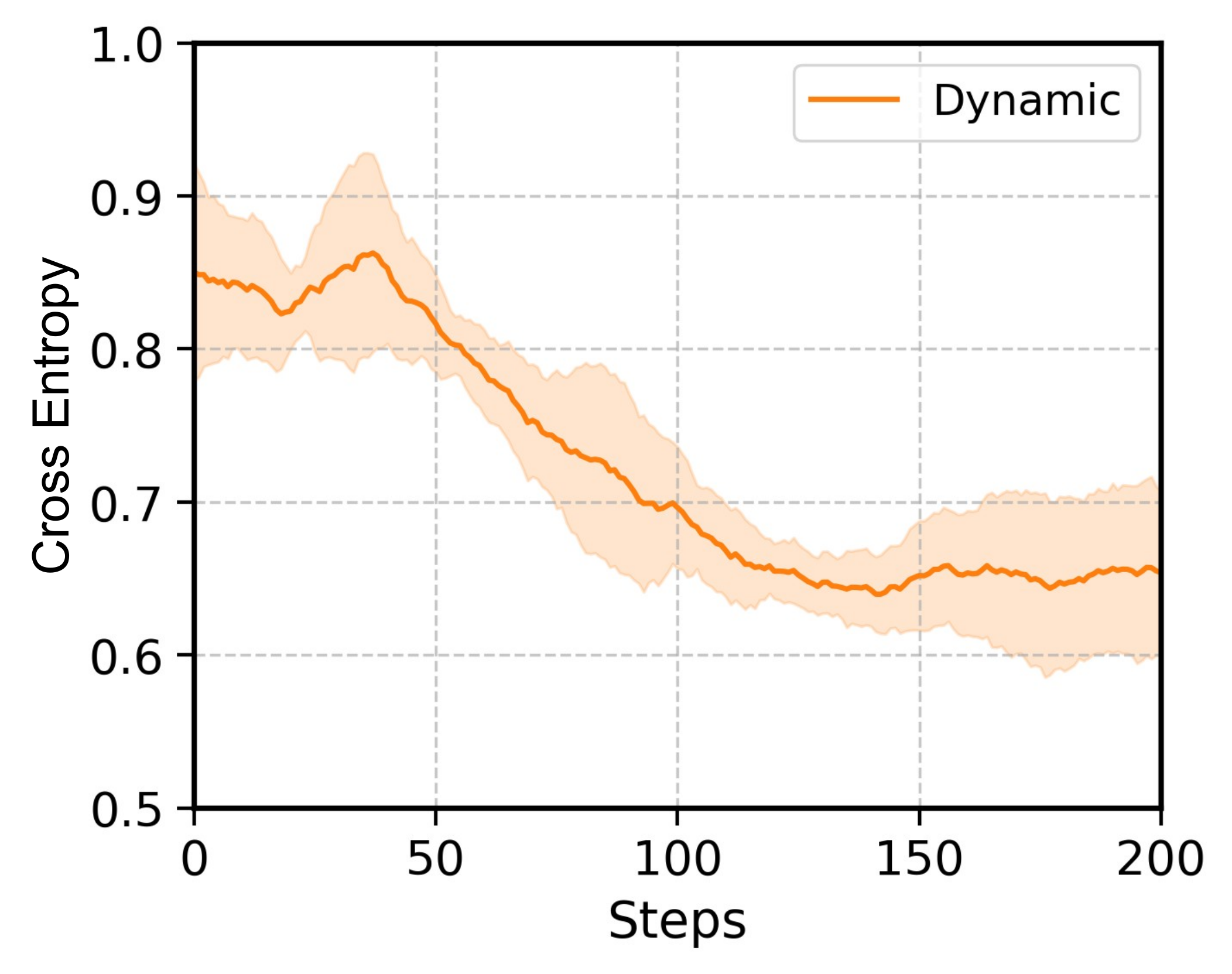}}  

  \caption{\textbf{Training dynamics on Prompt-Collection-v0.1 (top) and APPS (bottom).} 
  (a, d) \textit{Dynamic} (orange) achieves continuous reward growth, whereas \textit{Static} (blue) saturates and \textit{w/o constraint} (red) collapses.
  (b, e) The KL divergence of \textit{Dynamic} rises steadily, indicating deep exploration beyond the initial policy $\pi_0$, while \textit{Static} remains tethered.
  (c, f) The cross entropy remains low, confirming that the Refiner $\pi_{\text{refiner}}$ successfully tracks the evolving policy $\pi_\theta$.}
  \label{fig:combined_training_curves}
\end{figure*}

\textbf{Reward and KL Dynamics.} Figure~\ref{fig:combined_training_curves} reveals a fundamental difference in optimization trajectories. The reward curve for the Dynamic method increases steadily alongside a monotonically increasing KL divergence. This confirms that \(\pi_\theta\) continuously explores new regions of the parameter space, unhindered by a fixed anchor. In contrast, the Static baseline sees its KL divergence plateau early, suggesting the policy has hit the boundary of the trust region defined by \(\pi_0\), effectively halting further exploration. The w/o Constraint baseline initially improves but eventually suffers from distribution collapse, evidenced by a sudden drop in reward and a spike in KL divergence. This result validates our hypothesis that dynamic constraints enable sustained exploration where static constraints saturate and unconstrained RL collapses.

\textbf{Online Refiner Stability.} The stability of the cross-entropy loss serves as a proxy metric for the refiner's reliability. Despite the policy \(\pi_\theta\) drifting significantly from its initialization (high KL), the distance between \(\pi_\theta\) and \(\pi_{\text{refiner}}\) (cross entropy) remains low and stable. This implies that the refiner successfully acts as a "moving anchor," adapting to the policy's evolution while maintaining local regularization. The refiner effectively approximates a gold model, guiding the policy through the optimization landscape without tethering it to the starting point.

\subsection{Comparing to DAPO on Code Generation}
\label{sec:exp_part2}
\textbf{Setup.} The analysis of training dynamics mainly serves to understand the mechanism of dynamic constraint. To further compare the practical performance of our method against state-of-the-art approach, we select DAPO as the baseline, and implement the dynamic constraint based on DAPO by using it as a penalty like \citet{ouyang2022training}. Implementation details for both methods are provided in Appendix~\ref{appendix:hyperparameter}.

We employ RFT on the training set of APPS, and use the same verifiable reward function as in Sec.~\ref{sec:exp_part1}. After training, we evaluate both methods on four Python code-generation benchmarks: HumanEval, HumanEval+, MBPP, and MBPP++ and report \textsc{Pass@1} using bigcode-evaluation-harness\footnote{\url{https://github.com/bigcode-project/bigcode-evaluation-harness}}. In practice, RFT for code generation typically begins with a strong checkpoint to maximize performance gains. Accordingly, we adopt Qwen2.5-Coder-1.5B-Instruct\footnote{\url{https://huggingface.co/Qwen/Qwen2.5-Coder-1.5B-Instruct}}, a model trained on large-scale dataset that includes extensive code-centric corpora. All methods perform fine-tuning based on Qwen2.5-Coder-1.5B-Instruct.

\begin{figure*}[t!hb]
  \centering
  \subfigure[Task reward\label{fig:qwen_coder_apps}]{\includegraphics[width=0.38\linewidth]{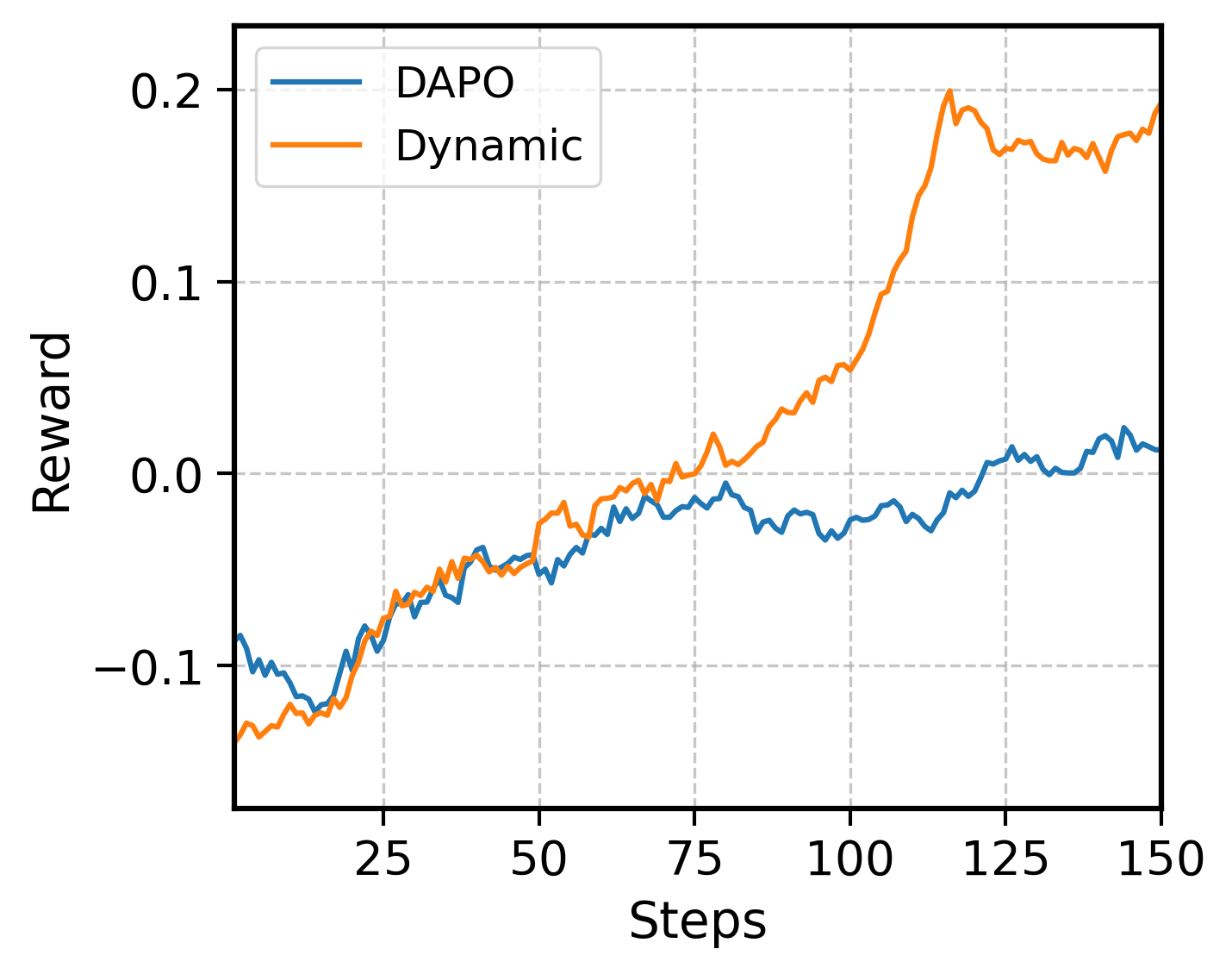}}
  \subfigure[Improvement ratio of Refiner LLM\label{fig:apps_ratio}]{\includegraphics[width=0.39\linewidth]{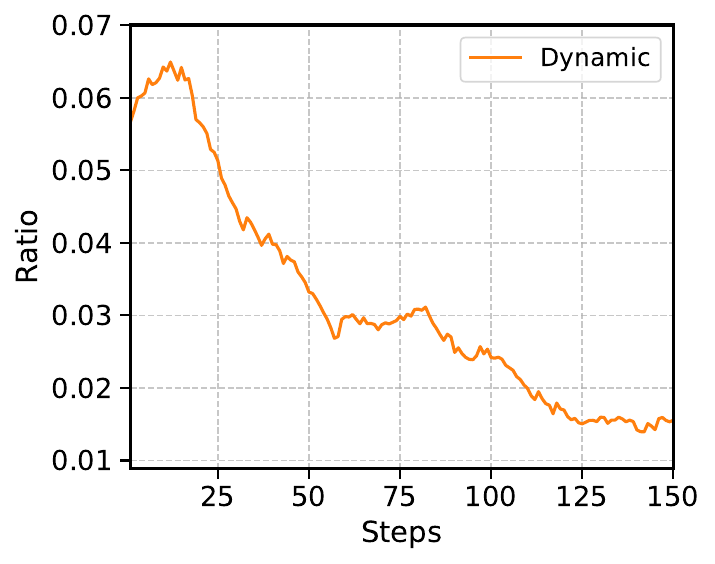}}
    
  \caption{\textbf{Training curves compared with DAPO on the APPS dataset.} The right figure shows the proportion of rollouts improved by the Refiner LLM.}
  \label{fig:dapo_curves}
\end{figure*}

\begin{table*}[thpt]
\centering
\small
\setlength{\tabcolsep}{1.5pt}
\begin{tabular}{lccccc}
\toprule
\textbf{Model} & \textbf{HumanEval} & \textbf{HumanEval+} & \textbf{MBPP} & \textbf{MBPP+} & \textbf{Avg.} \\
\midrule
Qwen2.5-Coder-1.5B-Instruct & 12.8 & 11.6 & 36.2 & 40.2 & 25.2 \\
DAPO    & 18.0 (+40.6\%) & 15.9 (+37.1\%) & 39.4 (+8.8\%) & 43.9 (+9.2\%) & 29.3 (+16.3\%) \\
Dynamic & 20.7 (+61.7\%) & 19.9 (+71.6\%) & 44.4 (+22.7\%) & 46.8 (+16.4\%) & 33.0 (+30.8\%) \\
\bottomrule
\end{tabular}
\caption{\textbf{Pass@1 (\%) on Python code generation benchmarks.} 
Qwen2.5-Coder-1.5B-Instruct is fine-tuned using DAPO and our method, respectively, 
and evaluated on four code generation benchmarks with the \texttt{bigcode-evaluation-harness}.}

\label{tab:benchmarks}
\end{table*}

\begin{figure*}[t!hbp]
    \centering
    \subfigure[Original response\label{fig:orig}]{
        \fbox{%
            \includegraphics[width=0.3915\textwidth]{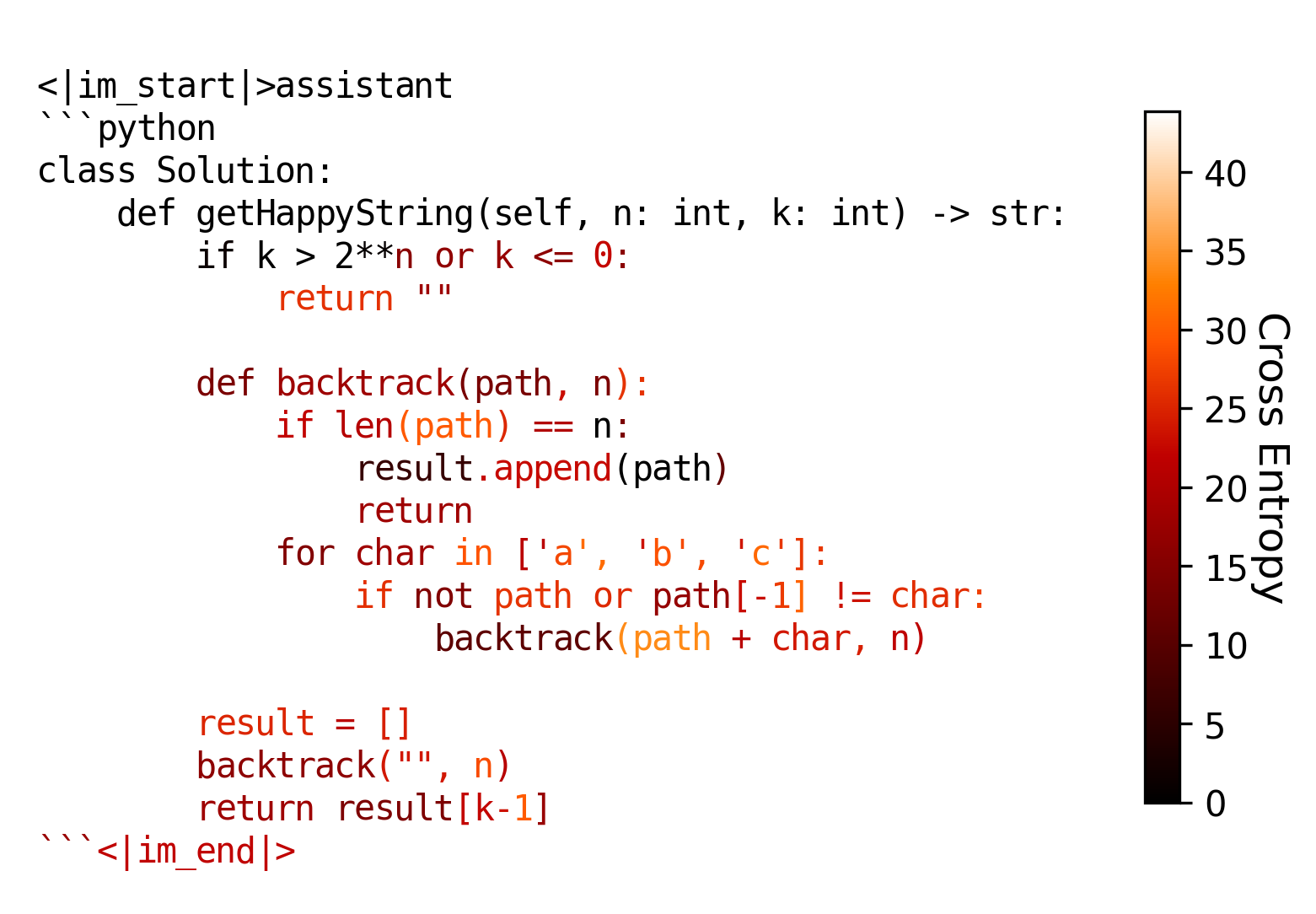}%
        }
    }
    \subfigure[Refined response\label{fig:refine}]{
        \fbox{%
            \includegraphics[width=0.4077\textwidth]{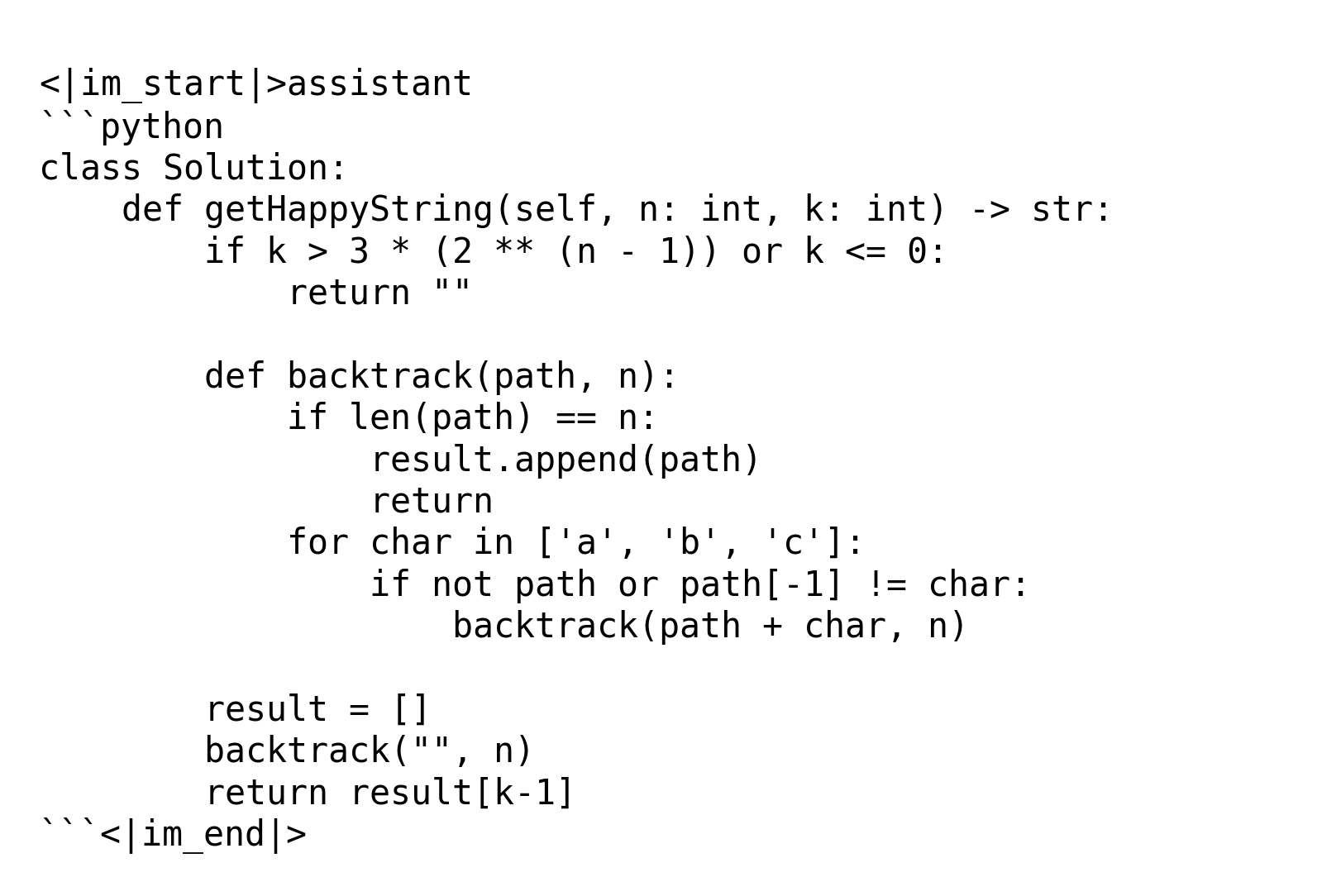}%
        }
    }
    \caption{\textbf{Demonstration of how the dynamic constraint refines model responses.} The two panels show Python backtracking solutions for the LeetCode problem "1415. The k-th Lexicographical String of All Happy Strings of Length n".
    (a) The left panel presents the original response, which contains an indexing error.
    (b) The right panel shows the refined response produced by the Refiner LLM, where the model enforces a more accurate condition (\texttt{3*(2*(n-1))}).
    The color on the left indicates the cross entropy values for each token.}

    \label{fig:qualitive_analysis}
\end{figure*}

\textbf{Benchmarking Results.} As shown in Figure~\ref{fig:qwen_coder_apps}, \textit{Dynamic} achieves a significantly higher asymptotic reward than DAPO. Notably, the reward curve exhibits a sharp acceleration around step 75, suggesting the discovery of novel optimization pathways that were likely suppressed by static regularization.
Table~\ref{tab:benchmarks} confirms that this advantage generalizes. \textit{Dynamic} outperforms DAPO by a significant margin (+30.8\% relative gain on average) across all test sets. This result suggests that dynamic constraints not only improve in-domain optimization but also preserve general coding capabilities better than static regularization.

\textbf{Adaptive Intervention.} Figure~\ref{fig:apps_ratio} plots the ``improvement ratio'' indicating the fraction of rollouts where the Refiner yields a higher reward than the policy $\pi_\theta$. This ratio naturally decays as the policy improves, confirming that our constraint is self-annealing. The dynamic constraint provides strong guidance early in training and relaxes as the policy becomes competent. Nevertheless, even when fewer than 10\% of rollouts are improved, the dynamic constraint still provides effective optimization guidance. Further ablation studies on the filter mechanism and dynamic constraint coefficient $\eta $ could be found in Appendix~\ref{appendix:ablation}.

\textbf{Qualitative Analysis.} Figure~\ref{fig:qualitive_analysis} visualizes the correction mechanism on a backtracking problem. The heatmap shows token-level cross-entropy. We observe that the constraint is sparse: for correct segments, the loss is near zero, activating only when the policy deviates into error. The Refiner successfully corrects a logical indexing error, guiding the policy from a failing state ($r=-0.3$) to a perfect solution ($r=1.0$). This ``intervention-on-demand'' behavior contrasts sharply with static KL, which indiscriminately penalizes deviation. Additional examples are provided in Appendix~\ref{appendix:qualitative}.

\textbf{Computational Cost Analysis.} The introduction of the online refiner tends to result in additional inference overhead. We compare the training time of DAPO and our Dynamic method using 4 NVIDIA A6000 GPUs. The total training time for DAPO is approximately 17.8 hours, while the Dynamic method requires 26.3 hours. This corresponds to a 48\% increase in training time. However, given the substantial performance improvement observed in benchmarks, we consider this trade-off to be reasonable in many cases.

\textbf{Qualitative Analysis.} Figure~\ref{fig:qualitive_analysis} presents a detailed example illustrating the effect of the dynamic constraint. We extract an original response and its corresponding refined response during training, and compute the token-wise cross entropy to analyze the constraint at the token level. The original response attains a reward of \(-0.3\) (compile passes but unit test fails), whereas the refined response achieves a reward of \(1.0\), successfully correcting the errors in the original response. The refined response repeats the original content in the initial segments, resulting in overlapping tokens with cross entropy values close to zero. After the point of modification, the cross entropy rises sharply and remains high thereafter. The dynamic constraint applies to the entire sequence following the modification, since in a causal language model, changing a single token affects all subsequent tokens. Therefore, aligning the entire remaining sequence to the refined response is reasonable. Additional examples are provided in Appendix~\ref{appendix:qualitative}.

\section{Related Work}

\textbf{KL Regularization in RFT.}  
The earliest work by \citet{stiennon2020learning} introduces KL regularization to constrain the magnitude of policy updates during fine-tuning. This approach addresses the out-of-distribution (OOD) issue that arises because reward models are typically trained on outputs generated by the reference model and annotated by humans. InstructGPT \citep{ouyang2022training} adopts the same mechanism for similar reasons, and subsequent models such as Kimi K1.5 \citep{team2025kimi} and DeepSeek-R1 \citep{guo2025deepseek} continue this practice for training reasoning models. Following DAPO~\citep{yu2025dapo}, KL regularization has been gradually removed from RFT algorithms as it restricts model performance. This aligns with the finding in \citet{bai2022training} that $\sqrt{D_{\mathrm{KL}}(\pi_\theta \| \pi_0)}$ correlates approximately linearly with task reward. However, ProRL~\citep{liu2025prorl} demonstrates that KL regularization can help balance exploration and exploitation in the policy. Our work seeks to relax the KL divergence constraint to the greatest extent possible, striking a balance that preserves the stability it provides while preventing it from stifling the model's exploratory potential.


\textbf{Hybrid RFT and SFT.}
To bridge the gap between the imitation limits of SFT \citep{shenfeld2025rl, kirk2023understanding} and the inherent instability of RFT \citep{lv2025unified}, hybrid approaches like UFT~\citep{liu2025uft} and SRFT~\citep{fu2025srft} jointly optimize these objectives to balance imitation and exploration. Recent work further integrates them via prefix-conditioning~\citep{huang2025prefixrft} or unified theoretical frameworks~\citep{lv2025unified}. However, these methods typically derive the supervised signal from static offline datasets, causing the SFT loss to act as a fixed regularization term that pulls the policy toward potentially suboptimal demonstrations. Our approach departs from this by introducing a dynamic constraint that adaptively refines policy rollouts to provide an evolving supervised signal. This online mechanism stabilizes training without tethering the policy to an outdated reference, thereby preserving exploration efficiency while constraining unsafe drift.


\section{Conclusion}
The fundamental limitation of KL regularization lies in its static nature. By anchoring the policy to a fixed reference, static constraints create a trade-off between stability and optimality. We address this by introducing dynamic constraints that evolve with the policy, transforming the reference from a fixed anchor into an adaptive guide.

Our key insight is that constraints should intervene only when necessary. By using the reference model as an online refiner that provides minimal corrections, the constraint automatically strengthens when outputs degrade and relaxes when they improve. This enables policies to explore beyond the pretrained distribution while maintaining language coherence.

Empirical results across dialogue and code generation tasks demonstrate that dynamic constraints resolve the KL regularization dilemma. Policies achieve substantially higher rewards while maintaining stability, even as they diverge significantly from the reference model. Our work suggests that the future of reinforcement learning fine-tuning may lie not in removing constraints, but in making them adaptive.

\clearpage

\bibliography{adappo}

\clearpage

\appendix

\section{Experiments Details}
\label{appendix:exp_details}
\subsection{Hyperparameters}
\label{appendix:hyperparameter}

We summarize the hyperparameter settings used in Table~\ref{table:hyperparam}. The exmperiment codes are based on \texttt{OpenRLHF}\footnote{\url{https://github.com/OpenRLHF/OpenRLHF}}. Most hyperparameters follow the default configurations provided by the respective codebases. We adjust the batch size according to available computational resources, and focus primarily on tuning the learning rate and KL coefficient.

We adopt a sequential tuning strategy: we first search for an appropriate learning rate, then tune the KL coefficient. Intentionally, we increase the learning rate to the point where, under a default KL coefficient, the PPO training with a static constraint begins to exhibit distributional collapse. This setup is designed to satisfy our core hypothesis (illustrated in Figure~\ref{fig:graph_abs}): once $\pi_\theta$ moves beyond the “safe region” around $\pi_0$, the static constraint hinders further updates, while the dynamic constraint does not. After identifying this collapse point, we gradually refine the KL coefficient until the static constraint is able to maintain a balance between stable training and performance. Then, the same hyperparameters are directly applied on the dynamic constraint.

Our experimentation employs 2 AMD EPYC 7773X CPUs and 8 NVIDIA A6000 GPUs (48GB each).

\begin{table}[h!tp]
\caption{Hyperparameters in Section~\ref{sec:exp_part1}.}
\centering
\renewcommand{\arraystretch}{1.2}
\begin{tabularx}{0.9\linewidth}{l *{2}{>{\centering\arraybackslash}X}}
\toprule
\textbf{Hyperparameter} & \textbf{Prompt-Collection-v0.1} & \textbf{APPS} \\
\midrule
Actor Learning Rate & 1e-6 & 1e-6 \\
Critic Learning Rate & 9e-6 & 9e-6 \\
Epochs & 1 & 7 \\
PPO Epoch & 1 & 4 \\
Batch Size & 64 & 128 \\
Mini Batch Size & 16 & 16 \\
Gradient Accum. Steps & 4 & 8 \\
Iterations & 430 & 200 \\
KL Coefficient ($\eta$) & 1e-2 & 1e-4 \\
Discount ($\gamma$) & 1 & 1 \\
GAE ($\lambda$) & 0.95 & 0.95 \\
Clip Ratio ($\epsilon$) & 0.2 & 0.2 \\
Value Clip Range & 0.2 & 0.2 \\
\bottomrule
\end{tabularx}
\label{table:hyperparam}
\end{table}

\begin{table}[tbp]
\caption{Hyperparameters in Section~\ref{sec:exp_part2}.}
\centering
\renewcommand{\arraystretch}{1.2}
\begin{tabularx}{0.9\linewidth}{l *{2}{>{\centering\arraybackslash}X}}
\toprule
\textbf{Hyperparameter} & \textbf{DAPO} & \textbf{Dynamic} \\
\midrule
Actor Learning Rate & 1e-6 & 1e-6 \\
Epochs & 5 & 5 \\
Batch Size & 64 & 64 \\
Mini Batch Size & 8 & 8 \\
Samples per Prompt & 8 & 8 \\
Iterations & 150 & 150 \\
KL Coefficient ($\eta$) & 1e-2 & 1e-4 \\
Discount ($\gamma$) & 1 & 1 \\
Clip Ratio ($\epsilon_\mathrm{low}, \epsilon_\mathrm{high}$) & (0.2, 0.3) & (0.2, 0.3) \\
Value Clip Range & 0.5 & 0.5 \\
\bottomrule
\end{tabularx}
\label{table:hyperparam_exp2}
\end{table}

\subsection{Reward Setting for Code Generation}
We implement a two-tiered reward mechanism combining coarse-grained feedback and adaptive testing incentives. The coarse-grained reward follows the same setting as \citep{le2022coderl} :
\label{appendix:reward}
\begin{equation}
R_{\text{coarse}}(\hat{W}) = \begin{cases}
-1.0, & FB(\hat{W}) \text{ is syntax error} \\
-0.6, & FB(\hat{W}) \text{ is runtime error} \\
-0.3, & FB(\hat{W}) \text{ is unit test failure} \\
1.0, & FB(\hat{W}) \text{ is all pass}
\end{cases}
\end{equation}

Coarse-grained feedback serves as an incentive mechanism for language models, increasing the probability of generating correcte code reduceing the chances of producing erroneous code.
When the $FB(\hat{W})$ is between  pass and failure, we introduce adaptive feedback proportional to the test passing rate to address reward sparsity and encourage maximal test coverage.
\begin{equation}
    R_{\text{adaptive}}(\hat{W}) = -0.3 + 1.3 \times \frac{N_{\text{pass}}}{N_{\text{pass}} + N_{\text{fail}}}
\end{equation}

Where $FB(\hat{W})$  represents the compiler feedback for generated code  $\hat{W}$,and the value of $R_{\text {adaptive }}(\hat{W})$ is determined by the pass rate of the unit test.

\subsection{Prompt Detail}
\label{appendix:prompt_detail}

The following prompt is used to instruct the Refiner LLM in the APPS code generation experiments. The prompt guides the refiner to either reproduce correct solutions verbatim or apply minimal corrections to incorrect ones.

\begin{tcolorbox}[
  colback=gray!3,
  colframe=gray!50,
  title={\textcolor{black}{\textbf{Refiner Prompt for APPS}}},
  fonttitle=\scriptsize\normalsize,
  boxrule=0.4pt,
  arc=3pt,
  left=4pt,
  right=4pt,
  top=4pt,
  bottom=4pt,
  fontupper=\scriptsize\small,
]
You are a coding assistant for refining program solutions.

You will receive:\\
- Question: A programming problem description.\\
- Response: A proposed code solution.\\
- Reward: A numerical score measuring correctness, defined as:\\
  \hspace{0.5em}- -1.0 $\to$ Compilation failed\\
  \hspace{0.5em}- -0.6 $\to$ Runtime error on test cases\\
  \hspace{0.5em}- $-0.3 + 1.3 \times \text{pass\_rate} \to$ Partial pass, where $\text{pass\_rate} \in [0, 1]$\\

\textbf{Your Task:}\\
1. If Reward = 1.0 (full correctness): Output the code from Response exactly, without changes.\\
2. If Reward < 1.0 (incorrect/incomplete):\\
   - Analyze Response to identify issues (syntax errors, runtime errors, logical mistakes, failing test cases).\\
   - Refine and fix the code to maximize correctness and improve reward.\\
   - Ensure the refined code follows the input/output format, covers edge cases, and is efficient enough.\\
3. Output Format: Only output the final code (either the same code if correct, or a refined version). Do not include additional explanations, comments, or extra text.\\

\textbf{Question:}
\{question\}

\textbf{Response:}
\{response\}

\textbf{Reward:} \{reward\}

\textbf{Refine:}
\end{tcolorbox}

\section{Ablation Studies}
\label{appendix:ablation}

We conduct ablation experiments to evaluate the effect of (1) the filtering mechanism described in Equation~\ref{eq:ce}, which ensures that only refined responses with higher rewards than the original responses are used to provide dynamic constraints, and (2) the dynamic constraint coefficient $\eta$, which controls the strength of the cross-entropy regularization term.

\begin{figure}[t!hb]
  \centering
  \subfigure[KL divergence]{\includegraphics[width=0.45\linewidth]{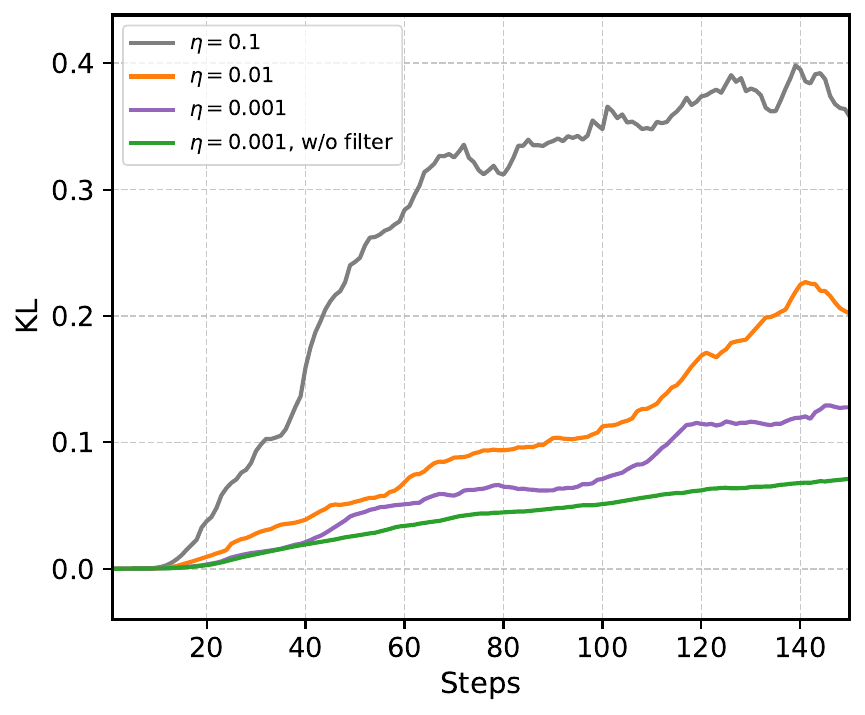}}
  \subfigure[Task reward]{\includegraphics[width=0.45\linewidth]{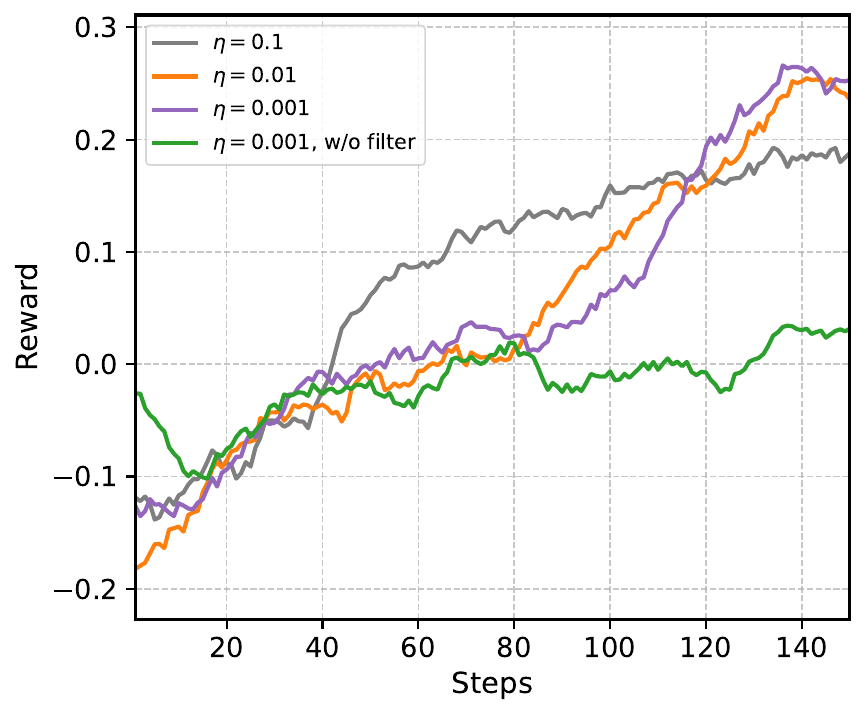}}

  \subfigure[Group reward std]{\includegraphics[width=0.45\linewidth]{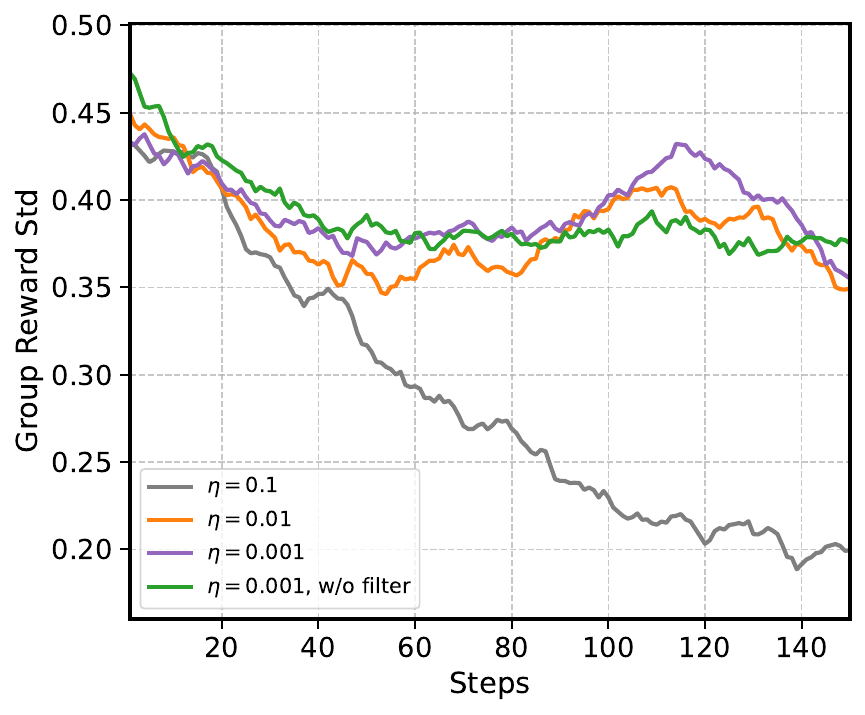}}
  \subfigure[Cross entropy]{\includegraphics[width=0.45\linewidth]{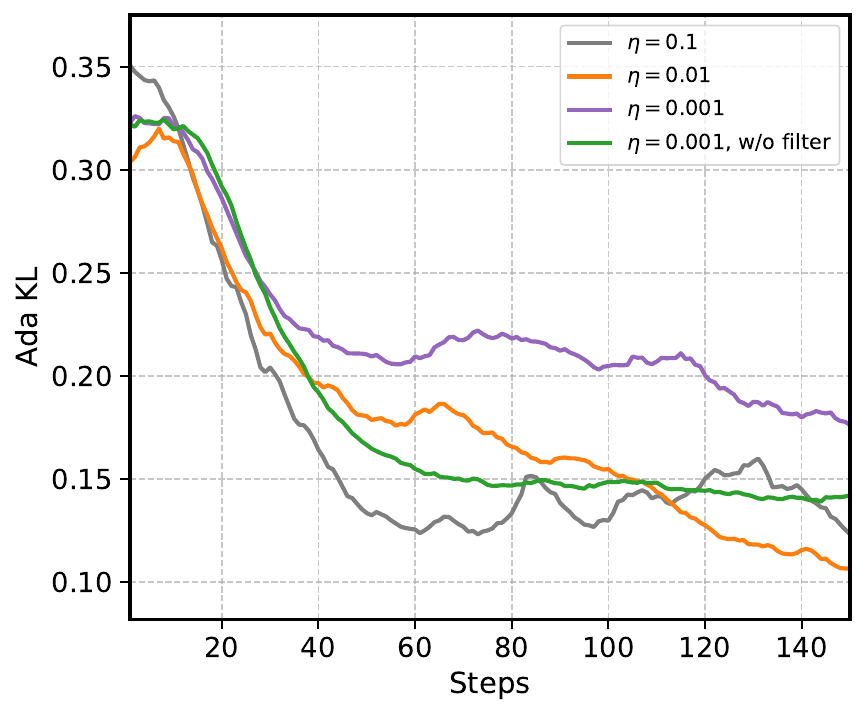}}
  \caption{\textbf{Ablation studies on the filtering mechanism and dynamic constraint coefficient $\eta$.} The curves show training dynamics under different configurations: with/without the filter that excludes refined responses with lower rewards, and with different values of $\eta$ that control the strength of the dynamic constraint.}
  \label{fig:ablation}
\end{figure}

\textbf{Effect of $\eta$.} As shown in Figure~\ref{fig:ablation}, a larger $\eta$ leads to rapid reward improvement in the early stages of training. However, over time, it negatively impacts the diversity of generated outputs, as evidenced by the declining group reward standard deviation. This reduced diversity ultimately limits the final achievable reward. Based on these observations, we select $\eta = 0.001$ as a balanced choice that maintains sufficient exploration while still benefiting from the dynamic constraint.

\textbf{Effect of Filter.} Without the filtering mechanism, the refiner occasionally produces responses with lower rewards than the original outputs from $\pi_\theta$. This phenomenon becomes more pronounced as $\pi_\theta$ improves during training, since stronger policies are harder to refine. Consequently, including such inferior refined responses degrades the quality of the dynamic constraint and hampers policy optimization. The filter effectively addresses this issue by excluding refinements that fail to improve upon the original response.

\section{Qualitative Analysis}
\label{appendix:qualitative}

\tcbset{
  table style/.style={
    colback=gray!5,
    colframe=gray!40,
    boxrule=0.4pt,
    arc=2pt,
    outer arc=2pt,
    boxsep=1pt,
  }
}

\begin{table*}[t!bh]
  \centering
  \renewcommand\arraystretch{1.3}
  \caption{Examples of Prompt-Collection-v0.1.}
  \label{table:prompt-collection-qualitative}

  \begin{tcolorbox}[table style, width=\textwidth]
    \begin{tabularx}{\textwidth}{@{}>{\hsize=.22\linewidth}X>{\hsize=.78\linewidth}X@{}}
      \toprule
      \textbf{Query $s_0$} & \textbf{Responses / Model Outputs} \\
      \midrule

      Name a kind of fungal hyphae that can live in outer space.
      & 
      \textbf{Response of $\pi_\theta$:} Armillaria ostoyae, a type of honey mushroom, has been found in space debris.\\
      & \textbf{Response of $\pi_0$:} Armillaria mellea.\\
      & \textbf{Response of $\pi_{\text{refiner}}$:} Armillaria ostoyae, a type of honey mushroom, has been found in space debris.\\
      \midrule

      Compute this: 7299729 + 114. Exclude words; show only the math.
      & 
      \textbf{Response of $\pi_\theta$:} 7299729 + 114 = 7299843\\
      & \textbf{Response of $\pi_0$:} 7299729 + 114 = 7299843\\
      & \textbf{Response of $\pi_{\text{refiner}}$:} 7299729 + 114 = 7299843\\
      \midrule

      What is the history of the Oxford dictionary? How did the dictionary come to be?
      &
      \textbf{Response of $\pi_\theta$:} Oxford Dictionary originated in 1884, initially a supplement to the Oxford English Dictionary (OED). First published 1884, updated by contributors, including...\\
      & \textbf{Response of $\pi_0$:} Founded in 1884 by James Murray, it evolved from a 14-volume set of dictionaries compiled by Samuel Johnson and others.\\
      & \textbf{Response of $\pi_{\text{refiner}}$:} Oxford Dictionary originated in 1884, initially a supplement to the Oxford English Dictionary (OED).\\
      \midrule

      Who is faster in real life, a tortoise or a hare?
      &
      \textbf{Response of $\pi_\theta$:} The tortoise.\\
      & \textbf{Response of $\pi_0$:} The tortoise.\\
      & \textbf{Response of $\pi_{\text{refiner}}$:} The hare.\\

      \bottomrule
    \end{tabularx}
  \end{tcolorbox}
\end{table*}

We provide a qualitative analysis on the \texttt{Prompt-Collection-v0.1} dataset in Table~\ref{table:prompt-collection-qualitative}. The first two examples show cases where $\pi_\theta$ produces accurate and appropriate answers. In these cases, $\pi_{\text{refiner}}$ stays close to $\pi_\theta$, while $\pi_0$ can drift and yield noticeably different outputs. The third example shows a truncation failure where $\pi_\theta$ hits the maximum generation length and ends with an incomplete sentence. $\pi_{\text{refiner}}$ detects this issue and keeps only the first complete sentence, removing the broken fragment. This example also highlights that when $\pi_0$ is far from $\pi_\theta$, it can act as a strong constraint even when $\pi_\theta$ is semantically correct. In the final example, both $\pi_\theta$ and $\pi_0$ give an incorrect answer, which may reflect biased knowledge from pretraining such as the tortoise and hare story. In contrast, $\pi_{\text{refiner}}$ corrects the error and returns the correct answer.

\begin{figure*}[thbp]
    \centering
    \subfigure[Original response]{
        \fbox{%
            \includegraphics[width=0.4\textwidth]{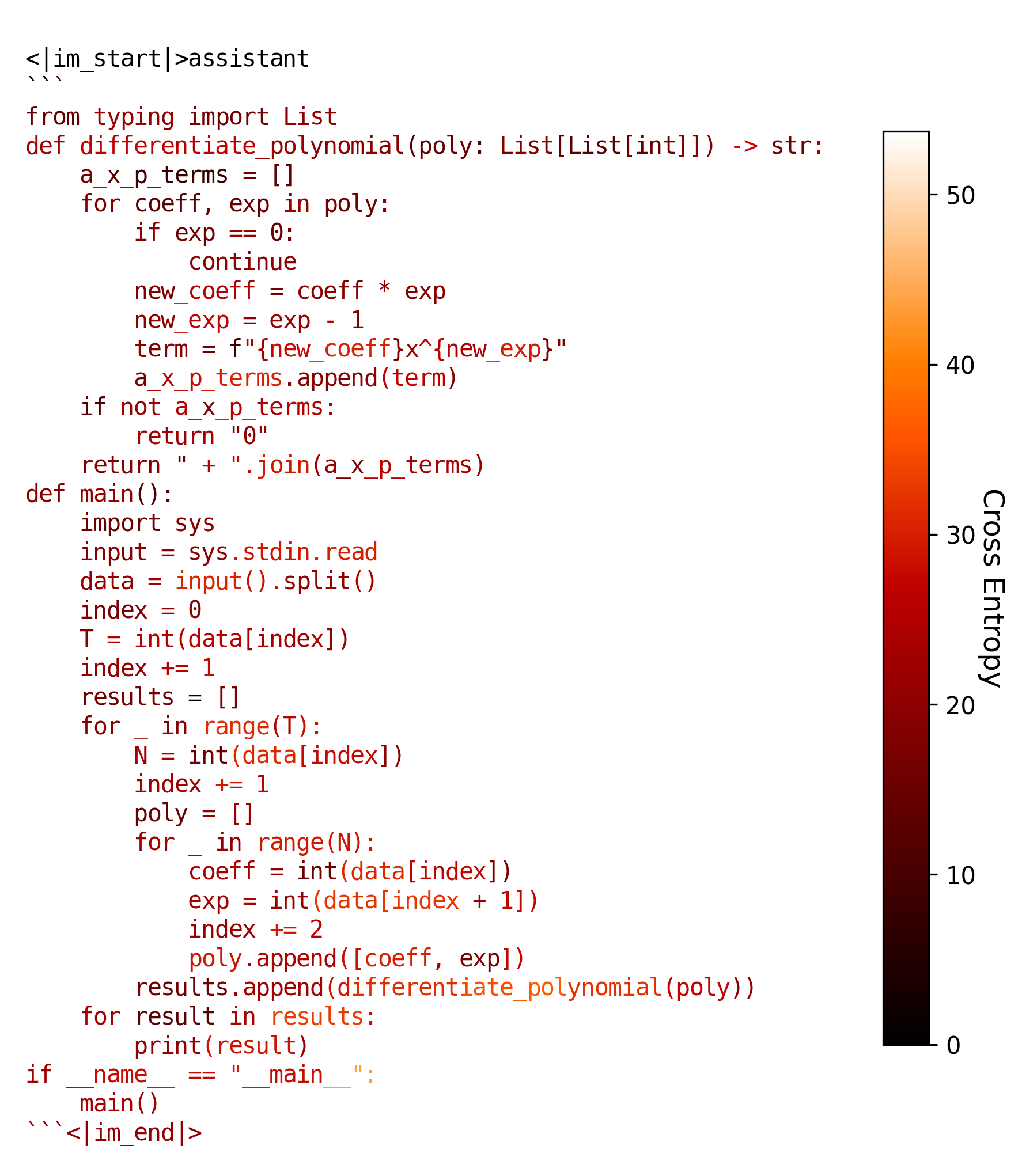}%
        }
    }
    \subfigure[Refined response]{
        \fbox{%
            \includegraphics[width=0.4\textwidth]{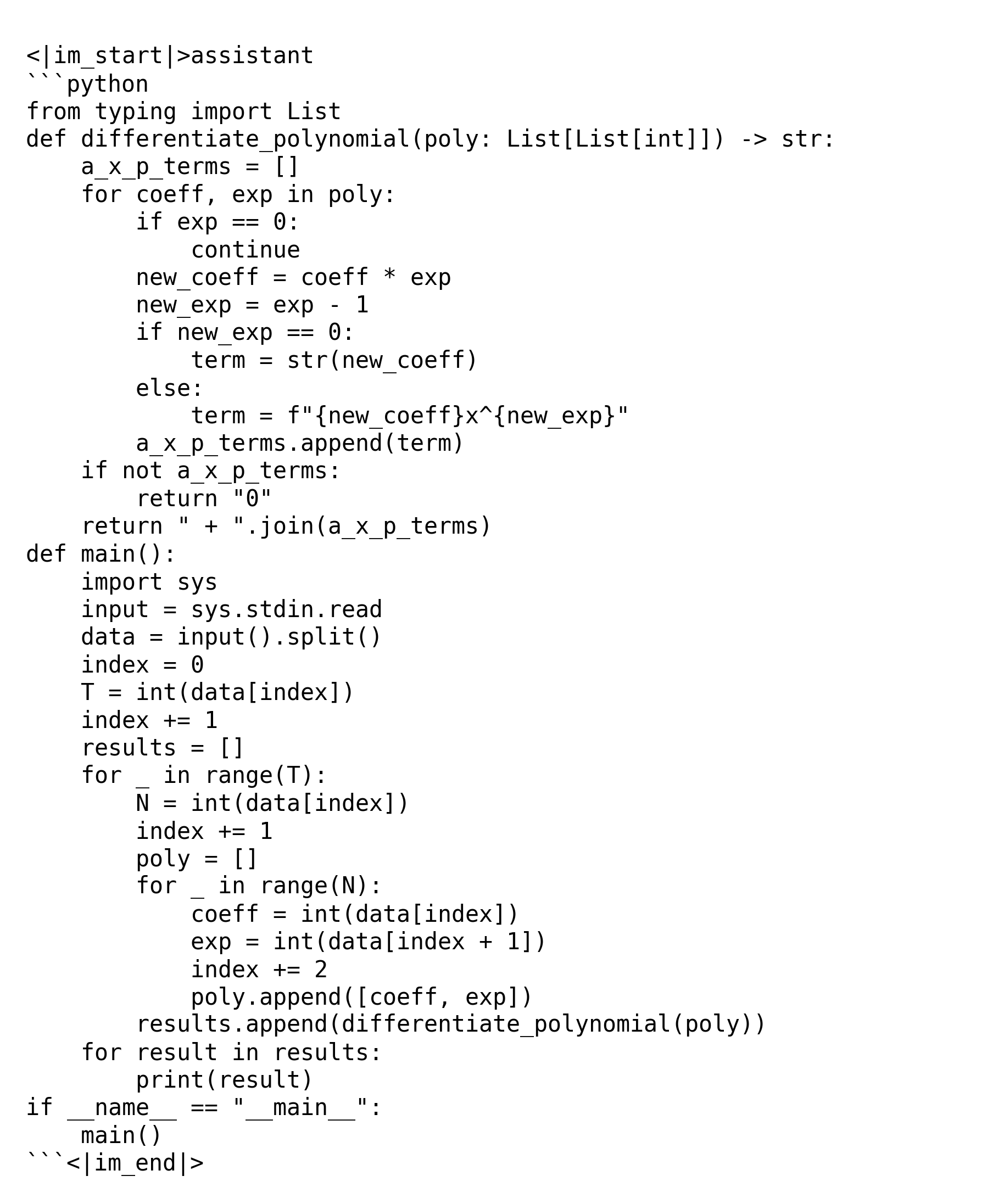}%
        }
    }
    \caption{The reward of the original response is $-0.3$, while that of the refined response is $1.0$. The original response also misses the initial signal \texttt{python}, which is correctly restored in the refined version.}
    \label{fig:apps-qualitative-13}
\end{figure*}

\begin{figure*}[thbp]
    \centering
    \subfigure[Original response]{
        \fbox{%
            \includegraphics[width=0.4\textwidth]{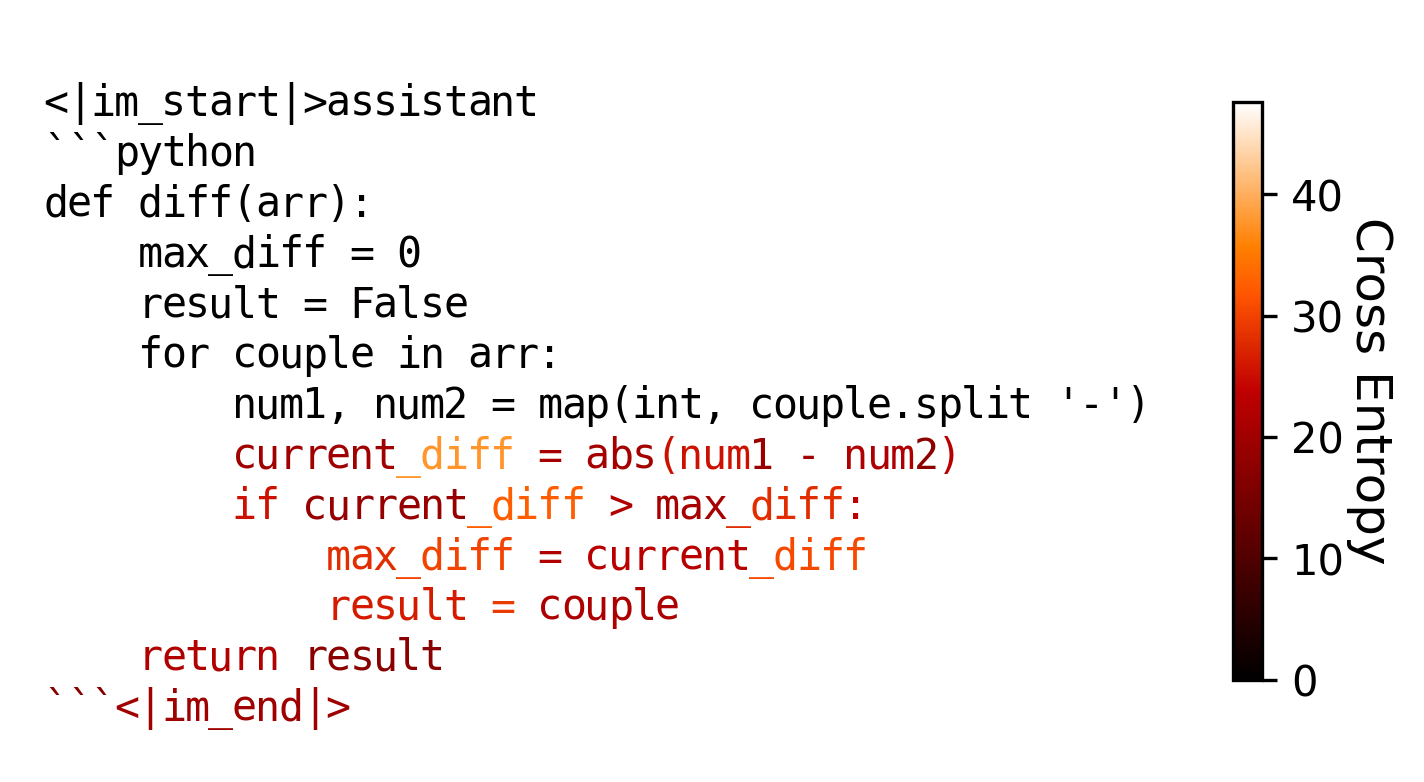}%
        }
    }
    \subfigure[Refined response]{
        \fbox{%
            \includegraphics[width=0.4\textwidth]{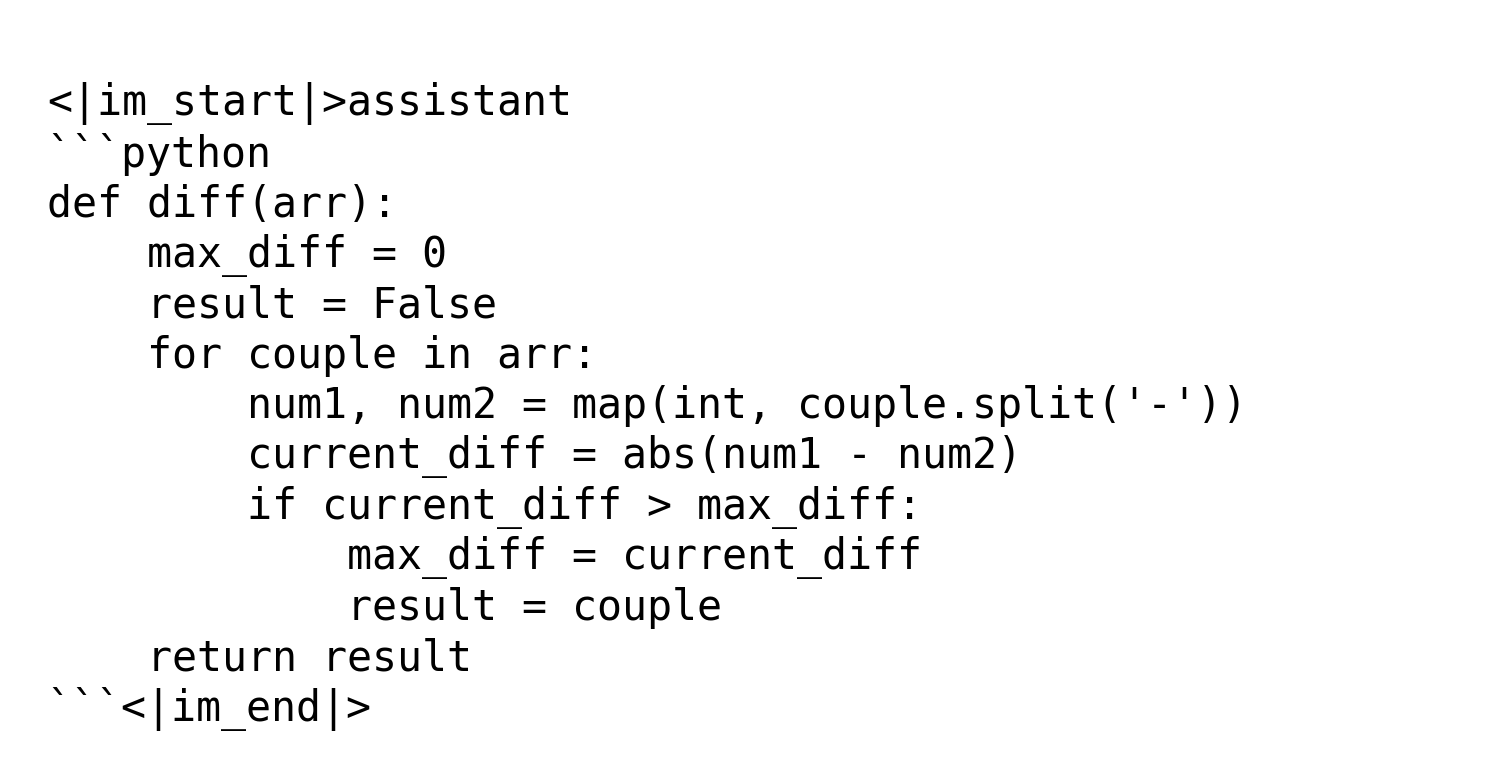}%
        }
    }
    \caption{The reward of the original response is $-1.0$, while that of the refined response is $1.0$. The refined response eliminates redundant operations and enforces a clearer logical structure.}
    \label{fig:apps-qualitative-68}
\end{figure*}

\begin{figure*}[thbp]
    \centering
    \subfigure[Original response]{
        \fbox{%
            \includegraphics[width=0.4\textwidth]{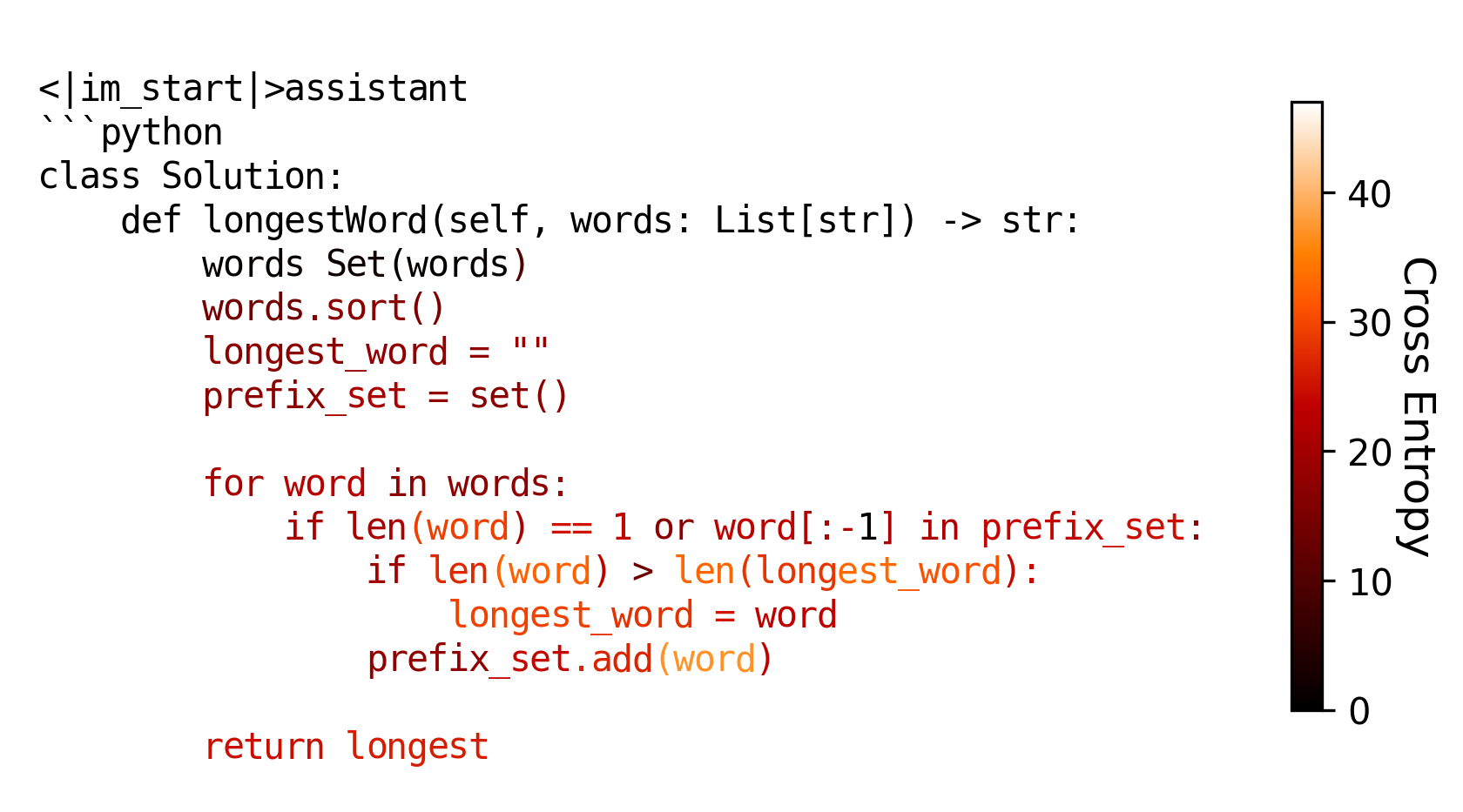}%
        }
    }
    \subfigure[Refined response]{
        \fbox{%
            \includegraphics[width=0.453\textwidth]{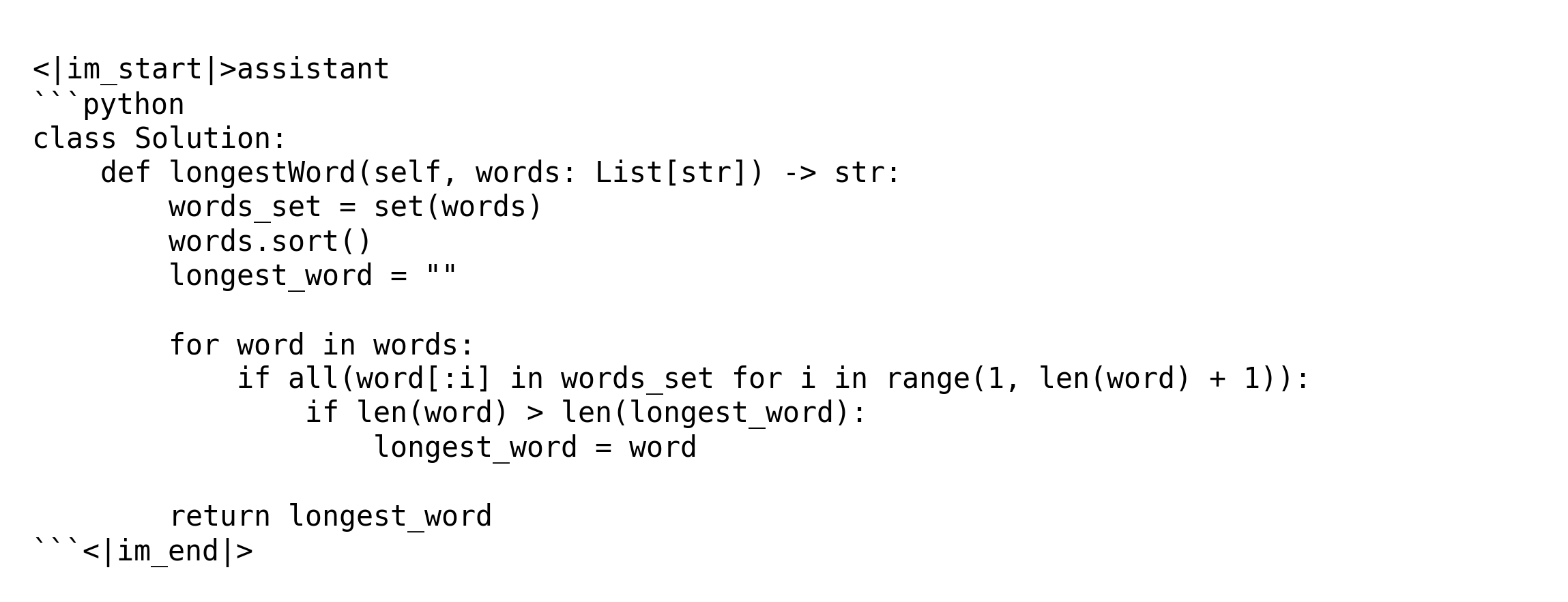}%
        }
    }
    \caption{The reward of the original response is $-1.0$, while that of the refined response is $1.0$. The original response produces an incorrect computation branch, which the refined response corrects by revising the conditional structure.}
    \label{fig:apps-qualitative-34}
\end{figure*}

\begin{figure*}[thbp]
    \centering
    \subfigure[Original response]{
        \fbox{%
            \includegraphics[width=0.453\textwidth]{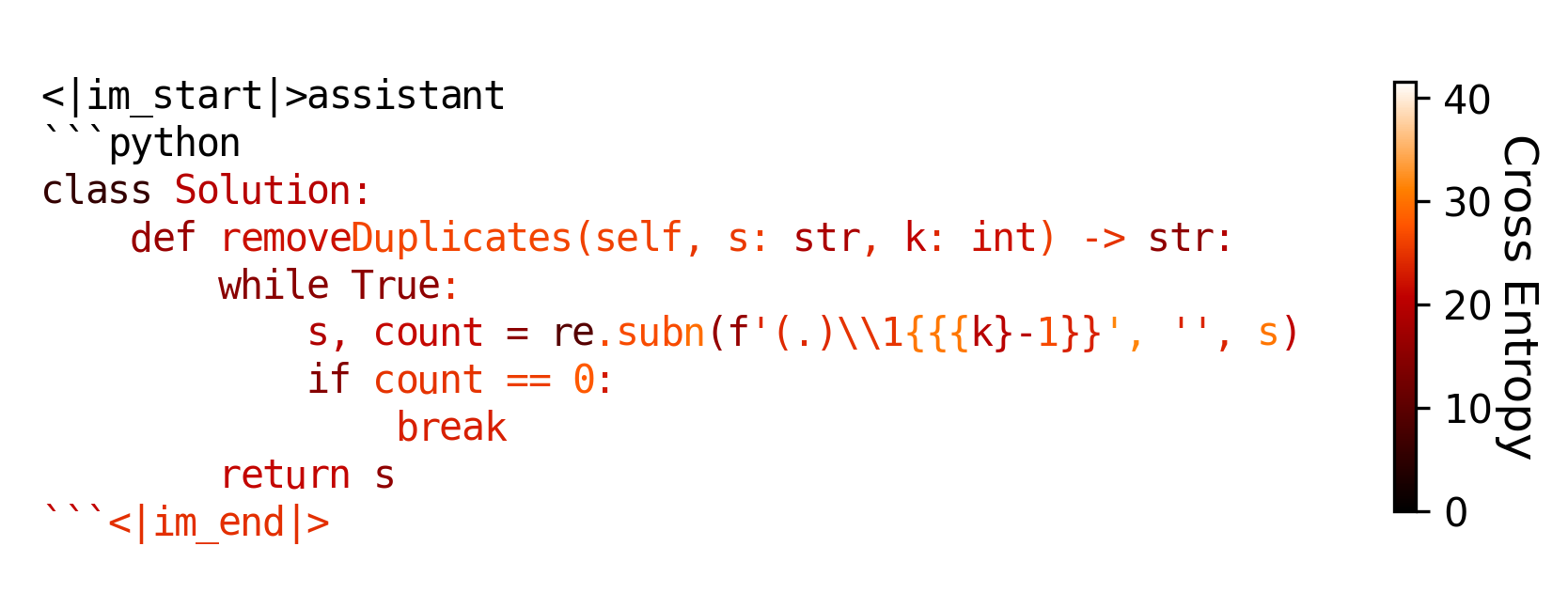}%
        }
    }
    \subfigure[Refined response]{
        \fbox{%
            \includegraphics[width=0.453\textwidth]{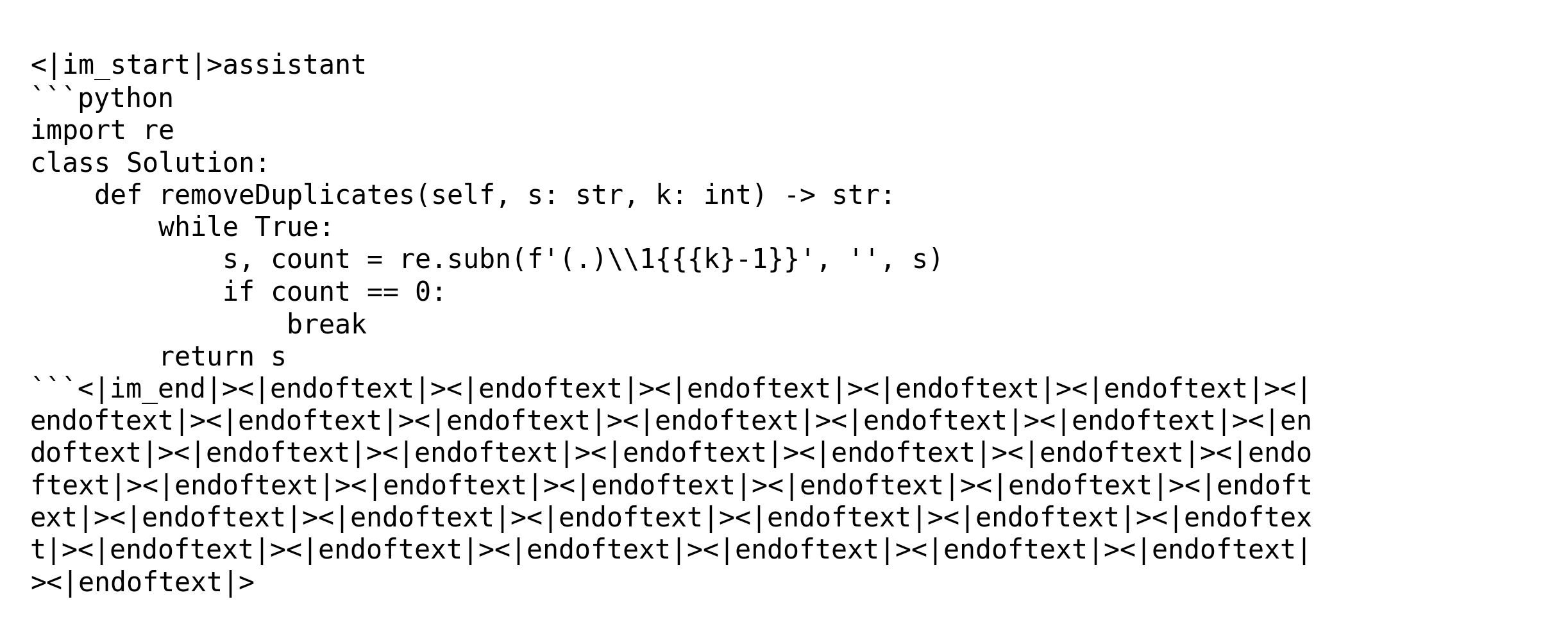}%
        }
    }
    \caption{The reward of the original response is $-0.6$, while that of the refined response is $1.0$. The refined version improves loop control and resolves syntax inconsistencies present in the original code.}
    \label{fig:apps-qualitative-41}
\end{figure*}

We also include qualitative examples on the APPS dataset in Figures~\ref{fig:apps-qualitative-13}, \ref{fig:apps-qualitative-68}, \ref{fig:apps-qualitative-34}, and \ref{fig:apps-qualitative-41}. Across these four cases, refinement consistently improves the reward from a negative value to $1.0$. The refiner restores missing format signals such as the \texttt{python} header, removes redundant operations, and makes the logic easier to follow. It also fixes wrong computation branches by rewriting conditional structures, and it repairs loop control and syntax issues that can cause evaluation failures.

\setlength{\subfigtopskip}{0pt}
\setlength{\subfigbottomskip}{0pt}
\setlength{\subfigcapskip}{2pt}
\setlength{\abovecaptionskip}{3pt}
\setlength{\belowcaptionskip}{0pt}

\newpage
\section{RFT from the Mirror Learning Perspective}
\label{appendix:proof}

\begin{definition}
\label{def:drift}
A \textit{drift functional} $\mathfrak{D}$ is a map 
\begin{equation}
    \mathfrak{D}:\Pi\times\mathcal{S} \rightarrow \{ \mathfrak{D}_{\pi}(\cdot|s):
    \mathcal{P}(\mathcal{A}) \rightarrow  \mathbb{R} \}, \nonumber
\end{equation}
such that for all $s\in\mathcal{S}$, and $\pi, \bar{\pi}\in\Pi$, writing $\mathfrak{D}_{\pi}(\bar{\pi}|s)$ for $\mathfrak{D}_{\pi}\big(\bar{\pi}(\cdot|s)|s\big)$, the following conditions are met
\begin{enumerate}[itemsep=0pt, topsep=0pt]
    \item  \label{bullet:drif2}$\mathfrak{D}_{\pi}(\bar{\pi}|s)\geq \mathfrak{D}_{\pi}(\pi|s) = 0$ \textit{(nonnegativity)},
    \item  \label{bullet:drif3}$\mathfrak{D}_{\pi}(\bar{\pi}|s)$ has zero gradient\footnote{More precisely, all its G\^ateaux derivatives are zero.} with respect to $\bar{\pi}(\cdot|s)$, evaluated at $\bar{\pi}(\cdot|s)=\pi(\cdot|s)$ \textit{(zero gradient)}.
\end{enumerate}
\end{definition}

\begin{definition}
We say that $\mathcal{N}:\Pi\rightarrow\mathbb{P}(\Pi)$ is a \textit{neighbourhood operator}, where $\mathbb{P}(\Pi)$ is the power set of $\Pi$, if
\begin{enumerate}[itemsep=0pt, topsep=0pt]
    \item It is a continuous map \textit{(continuity)},
    \item Every $\mathcal{N}(\pi)$ is a compact set \textit{ (compactness)},
    \item  There exists a metric $\chi:\Pi\times \Pi\rightarrow\mathbb{R}$, such that $\forall \pi \in \Pi$, there exists $\zeta>0$, such that $\chi(\pi, \bar{\pi})\leq \zeta$ implies $\bar{\pi}\in\mathcal{N}(\pi)$ \textit{(closed ball)}.
\end{enumerate}
The \textit{trivial} neighbourhood operator is $\mathcal{N}\equiv \Pi$.
\end{definition}

The core theoretical insight of mirror learning is as follows: if the conditions specified in the definition are satisfied, and a better policy lies within the defined neighborhood $\mathcal{N}(\pi_{\text{old}})$, then applying the mirror learning update (as shown in Equation~\ref{eq:mirror_learning_update}) guarantees that the value function improves monotonically for all states $s$, i.e., $V\pi_{\text{new}}(s) \geq V\pi_{\text{old}}(s), \forall s \in S$. Since $V(s)$ is bounded, this monotonic improvement ensures that the policy converges to the optimal one.

Under the mirror learning framework, a static constraint such as the KL regularization can be interpreted as defining a fixed neighborhood \( \mathcal{N}(\pi_0) \). When the policy update is small, this constraint may have little effect. However, as the update magnitude increases, it may become increasingly difficult to find an improved policy within \( \mathcal{N}(\pi_0) \), thereby hindering further optimization.

In contrast, the dynamic constraint we propose replaces this static neighborhood with \( \mathcal{N}(\pi_{\text{refiner}}) \), which helps mitigate this limitation. Assuming that \( \pi_{\text{refiner}} \) can effectively track the evolution of \( \pi_\theta \), it becomes more likely that improved policies remain accessible within the neighborhood. This property supports continuous policy improvement and facilitates convergence to the optimal policy.


\section{Limitations}
\label{appendix:limit}
The efficacy of the dynamic constraint paradigm is intrinsically coupled with the refinement capabilities of the reference model. While our filtering mechanism ensures training stability by discarding failed corrections, it necessarily introduces a trade-off in constraint density and robustness. Furthermore, for complex reasoning tasks that demand extremely long output trajectories, the autoregressive nature of the refiner presents a non-trivial challenge in maintaining global structural coherence. We anticipate that these limitations will naturally recede as the frontier of model capacity continues to advance. In the future, we plan to extend this framework by introducing a block-wise refinement mechanism, which we believe will be essential for scaling dynamic constraints to the most demanding reasoning and agentic tasks.

\end{document}